\pdfoutput=1
\documentclass[11pt]{article}

\usepackage[preprint]{acl}

\usepackage{times}
\usepackage{latexsym}
\usepackage{amsmath} 
\usepackage[nointegrals]{wasysym}
\usepackage[T1]{fontenc}
\usepackage{booktabs}
\usepackage{multirow}
\usepackage[utf8]{inputenc}
\usepackage{makecell}
\usepackage{microtype}
\usepackage[most]{tcolorbox}
\usepackage{inconsolata}
\usepackage{enumitem}
\usepackage{graphicx}
\usepackage{tabularx}
\usepackage{longtable}
\usepackage{float}
\usepackage{caption} 
\usepackage{cuted}

\title{From Repetition to Recognition: \\ Inductive Discovery of Disinformation Narratives}

\author{
  \textbf{Max Upravitelev\textsuperscript{1,2}}, 
  \textbf{Veronika Solopova\textsuperscript{1,2}}, 
  \textbf{Jing Yang\textsuperscript{1,2,3}}, 
  \textbf{Charlott Jakob\textsuperscript{1,2,5}}, \\ 
  \textbf{Alexandra Tsiakalou\textsuperscript{1,2}}, 
  \textbf{Neda Foroutan\textsuperscript{1,2}}, 
  \textbf{and Vera Schmitt\textsuperscript{1,2,3,4,5}} \\
  \textsuperscript{1}Technische Universit\"{a}t Berlin \\
  \textsuperscript{2}German Research Center for Artificial Intelligence (DFKI) \\
  \textsuperscript{3}BIFOLD – Berlin Institute for the Foundations of Learning and Data \\
  \textsuperscript{4}Centre for European Research in Trusted AI (CERTAIN) \\
  \textsuperscript{5}Johannes Gutenberg-Universität Mainz \\
  {\small \textbf{Correspondence:} \href{mailto:max.upravitelev@tu-berlin.de}{max.upravitelev@tu-berlin.de}}
}

\begin{document}
\maketitle
\begin{abstract}

In disinformation datasets, narratives are often understood as recurring interpretive patterns that group texts under narrative labels. Recent work formalized narrative mining as inductively inferring narrative labels from corpora, but its evaluation stays tied to predefined taxonomies, a closed-world setting that cannot capture narratives absent from the reference labels. We introduce a three-tier evaluation framework for unsupervised narrative label generation: recovery (against a corpus's own taxonomy), mining (against external label sets), and discovery (without predefined labels). Applying it, we compare clustering-based and graph-community-based pipelines across seven disinformation datasets, with human validation of discovery on two. The two families are complementary under automated metrics, but in a corpus with two prominent topics, clustering can reduce one topic to 2\% of generated labels while graph-based pipelines stay balanced. Discovery validation also reveals many singletons (narrative labels derived from single claims, 30–62\% of graph outputs), which clustering cannot produce. Annotators confirm many as recognizable disinformation narratives, suggesting that in open-world discovery the repetition assumed by narrative mining may be recognized outside the corpus, not within it. We release human-validated narrative candidate labels for the Climate Obstruction and PolyNarrative datasets to support taxonomy development and dataset extension.

\end{abstract}

\section{Introduction}

Computational narrative analysis has increasingly relied on clustering methods \citep{hanley2024specious, Gerard2026, Ash_Gauthier_Widmer_2024}. The reliance on clustering fits the repeatability assumption prevalent in disinformation narrative mining research \citep{2025dinam}, stating that narratives are inferred by grouping recurring textual patterns across frequently occurring claims. In parallel, graph-based work represents narratives as networks of actors or events and their relations \citep{Tangherlini2020,Norambuena2021NarrativeMaps}.
Disinformation narrative taxonomies developed manually by domain experts echo both method families by providing hierarchical structures for repeating patterns across individual text instances. While they offer valuable foundations for automated disinformation narrative detection, they cannot be exhaustive by design. Narratives evolve over time as new variations emerge, motivating updates such as the recent revision of the CARDS taxonomy \citep{Coan2021,Coan2026}. Manually curated taxonomies are also shaped by the perspectives and backgrounds of their authors: PolyNarrative \citep{2025pn}, for instance, notes a stronger representation of Western institutions reflecting its analysts' backgrounds. 
 \begin{figure}[t]
    \centering
    \vspace*{22pt}
    \includegraphics[width=\columnwidth]{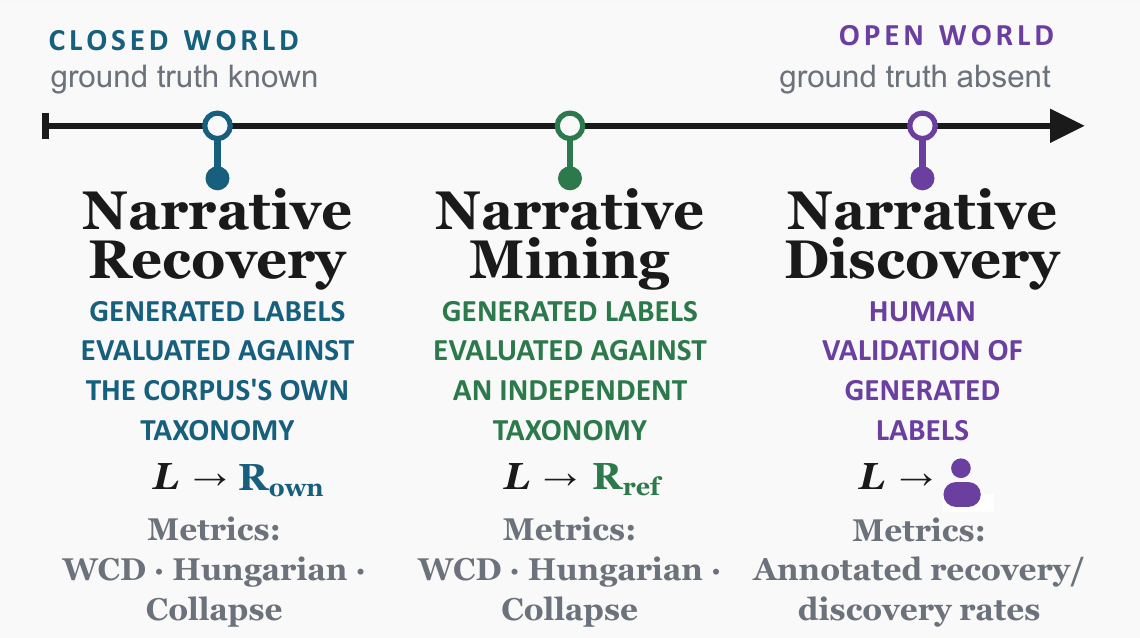}
    \caption{The proposed three-tier evaluation framework: narrative recovery, narrative mining, and narrative discovery. $L$: the labels a pipeline generates. $R_{\text{own}}$: the corpus's own taxonomy. $R_{\text{ref}}$: an independent reference taxonomy. Discovery has no reference set and is scored by the annotated yes/other/none rates of §\ref{sec:openworld}.}
     \label{fig:eval_framework}
 \end{figure}
Inductive narrative mining can help address these gaps by inferring narrative labels directly from corpora. The evaluation, however, still remains bounded by predefined taxonomies (e.g. DiNaM evaluated against \citealp{2024eu}). Even when the reference taxonomy comes from an independent dataset, evaluation is still limited to narratives already present in that label space. What is missing is a way to evaluate narratives that are novel relative to the reference taxonomy: cases absent from existing taxonomies that may nevertheless be valid candidates for new labels. This paper addresses that gap: narrative \emph{discovery}, approached with unsupervised methods that \emph{generate} narrative labels from a corpus instead of assigning texts to a fixed label set with classifiers. We also evaluate two closed-world settings (narrative recovery as a sanity check and narrative mining), since they enable automated evaluation where open-world discovery does not. We structure our study around the following research questions:  

\begin{itemize}
    \item \textbf{RQ1} How different are the narrative label sets produced by cluster-based and graph-based methods, and how do their outputs differ beyond what closed-world evaluation metrics capture?
    \item \textbf{RQ2} Does the within-corpus repeatability assumption established for narrative mining generalize to narrative discovery?
    \item  \textbf{RQ3} Can narrative mining be used to automatically identify novel narrative candidates that extend existing disinformation narrative taxonomies?
\end{itemize}

Our main contribution is a unified evaluation framework for unsupervised disinformation narrative label generation spanning closed- and open-world settings. It consists of three tiers (as illustrated in Fig.~\ref{fig:eval_framework}): recovery against a corpus's own taxonomy as a sanity check, mining against an independent reference taxonomy, and discovery without a taxonomy reference, validated by human annotators. The mining tier generalizes the cross-corpus evaluation of \citet{2025dinam} beyond fact-checking articles to arbitrary corpora, while recovery and discovery extend narrative mining to settings not previously formalized. We apply the framework to clustering- and graph-based inductive pipelines, two aggregation families prevalent in narrative-related work.
Across seven datasets covering several languages and topics, the two pipeline families are complementary under automated metrics, but several structural differences emerge. First, using clustering pipelines to generate labels at default parameters can suppress one of two main corpus topics, while graph pipelines preserve both topics across the parameter sweep. Second, the discovery tier identifies a long tail of singleton candidates (labels derived from a single claim) that clustering cannot produce, but which human annotators validate as disinformation narratives. We also found that LLM-as-judges perform differently from human validators, underlining the need for human-in-the-loop evaluation.
While the repeatability assumption has been definitional for disinformation narratives in the mining setting, we read our findings as pointing to a different role for repetition in narrative discovery: annotators appear to recognize singletons as candidates from what they already know, and not from patterns in the corpus at hand. To support future taxonomy maintenance, we release\footnote{\url{https://github.com/XplaiNLP/narrative-discovery}} human-validated narrative candidates for the Climate Obstruction and PolyNarrative datasets.

\section{Related Work and Preliminaries}
\label{sec:relatedwork}

Despite extensive research across narratology and computational approaches, there is no universally applicable definition of what constitutes a ``narrative'' \cite{piper-etal-2021-narrative}. Within disinformation datasets such as CARDS \citep{Coan2021}, EU DisinfoTest \citep{2024eu}, and PolyNarrative \citep{2025pn}, narratives are often understood as core messages. Following CARDS, we call such a core message a \emph{superclaim}: a normalized declarative assertion which a reader could agree or disagree with, and which is general enough to group multiple individual text instances under it. This is our criterion for counting a generated output as a narrative label and not as a single claim instance.

In recent years, many narrative-related approaches were built upon clustering mechanisms: \citet{hanley2024specious} 
cluster passage-level embeddings from unreliable news sites to identify and track narratives, \citet{Gerard2026} incorporate clustering in cross-platform narrative prediction, and Relatio \citep{Ash_Gauthier_Widmer_2024} clusters the entities of semantic role labeling (SRL) tuples to identify recurring relations between them in political discourse. A parallel line of work represents narratives as graphs of actors, events, and their relations, including actant networks \citep{Tangherlini2020} and AMR-based narrative signal graphs \citep{pournaki-willaert-2025}. 

Both families reappear in narrative mining, where respective groupings are summarized by LLMs, which generate a narrative label: claim-cluster pipelines \citep{2025dinam} and graph-based community summarization \citep{upravitelev-etal-2026-retrieving} adapted from NodeRAG \cite{xu2025noderagstructuringgraphbasedrag}. Both evaluate their outputs against pre-existing reference narratives.

Narrative discovery can be linked to a broader pattern in machine learning: category discovery from unlabeled data. Generalized category discovery \citep{vaze2022generalized} and open-world classification \citep{fei2016breaking} address the transition from closed- to open-world settings, but both rely on labeled known classes and score their handling of novel classes against ground-truth labels. The second requirement cannot be met in narrative discovery, where the candidates of interest are the ones no available taxonomy covers and which cannot be established in advance. Topic modeling requires no labeled classes. BERTopic \citep{2022bertopic} and LLM-based variants like TopicGPT \citep{pham-etal-2024-topicgpt} induce categories from unlabeled text directly, but these categories name subject areas instead of core messages: labels like \emph{Climate}, while narrative datasets require a full sentence like \emph{Climate solutions won't work}.

\section{Methodology}

We use \emph{narrative label} (NL) throughout for any label a pipeline generates, and \emph{narrative candidate} for an NL that human validation marked as a candidate for extending a taxonomy. App.~\ref{sec:terminology} collects the terminology.

\subsection{Evaluation Framework}

Our evaluation strategy organizes the assessment of inductive disinformation narrative-mining pipelines along a closed-to-open-world axis. The three tiers are conceptually related but independent. They are not meant as a sequential pipeline. Each tier is answering a distinct question about a pipeline's capability. The tiers are proposed on the assumption that success at one tier does not predict success at another: a pipeline can recover the source taxonomy well yet fail on mining, or produce strong mining results without surfacing discoveries.

\begin{enumerate}[nosep]
    \item \textbf{Narrative Recovery:} same corpus, its own taxonomy. This tier is a sanity check: can a pipeline infer a dataset's own taxonomy labels from that dataset's corpus? We propose re-purposing existing narrative classification datasets for this evaluation. Failure mode: A pipeline cannot reproduce the source taxonomy.
    \item \textbf{Narrative Mining:} same corpus, a different reference taxonomy. Can the pipeline's labels be matched to an independently developed taxonomy? Generalizes \citet{2025dinam}, beyond fact-checking corpora to arbitrary corpora containing disinformation. Failure mode: A pipeline scores well against the corpus's own taxonomy but poorly against an independent taxonomy from the same domain.
    \item \textbf{Narrative Discovery:} any corpus, no reference taxonomy. Can a pipeline identify narrative candidates not covered by any available taxonomy? Failure modes: annotators judge proposed candidates incoherent or already covered or a pipeline never proposes candidates annotators would have accepted.
\end{enumerate}

We argue that inductive narrative mining evaluation requires all three tiers: The closed-world tiers (recovery and mining) measure how closely the generated labels align with existing taxonomies, but cannot assess novel narrative candidates produced outside the reference label space. The discovery tier complements them: annotators judge the generated labels themselves, so no reference set is needed to score them.

\subsubsection{Closed-world metrics}
\label{sec:metrics}
Each pipeline produces $n$ NLs $L=\{l_1,\dots,l_n\}$, scored against a reference taxonomy $R=\{r_1,\dots,r_m\}$. Embedding both with harrier-oss-v1-0.6b \cite{microsoft_harrier_v1_2026} (chosen because its model family ranked first on the multilingual MTEB v2 \cite{enevoldsen2025mmteb} leaderboard\footnote{\url{https://huggingface.co/spaces/mteb/leaderboard}, accessed April 2026}) gives cosine similarity $S_{ij}=\cos(r_i,l_j)$, with $i\in[m]$ indexing references and $j\in[n]$ indexing NLs, and distance $D_{ij}=1-S_{ij}$. Throughout, $i$ and $j$ keep this convention. 

\paragraph{Hungarian similarity ($\uparrow$).} The mean cosine similarity of the best one-to-one alignment between $L$ and $R$: over alignments $\pi$ of size $k=\min(m,n)$ (sets of $k$ pairs $(i,j)$ in which each $r_i$ and each $l_j$ occurs at most once), take $\pi^{\star}=\arg\max_{\pi}\sum_{(i,j)\in\pi}S_{ij}$, computed via the Hungarian algorithm \citep{kuhn1955hungarian}, and report $\mathrm{Hungarian}(L,R)=\tfrac{1}{k}\sum_{(i,j)\in\pi^{\star}}S_{ij}$. Related to CEAF$_e$ \citep{luo2005coreference} and to the
cluster-class accuracy used in generalized category discovery \citep{vaze2022generalized}.

\paragraph{WCD: Weighted Chamfer Distance ($\downarrow$).}
$\mathrm{WCD}=\tfrac{1}{m+n}\bigl(\sum_{i}\min_{j}D_{ij}+\sum_{j}\min_{i}D_{ij}\bigr)$, combining reference coverage (left term) and NL precision (right term). Introduced for narrative mining by \citet{2025dinam}.

\paragraph{Collapse and collapse rate ($\downarrow$).}
Map each reference to its most similar NL, $\phi(i)=\arg\max_{j}S_{ij}$, and let $\mathrm{Collapse}=m-\lvert\{\phi(i):i\in[m]\}\rvert$, the number of references minus the number of distinct NLs they map onto, i.e. the number of top-1 collisions. It measures how well the NL set distinguishes references: if several references share one top-1 NL, that NL does not distinguish them. If three references all have the same NL as their nearest match, Collapse counts 2. Hungarian enforces a one-to-one matching and WCD averages over nearest neighbors, so neither reports these collisions. The rate $\mathrm{C/R}=\mathrm{Collapse}/m\in[0,(m{-}1)/m]$ normalizes for cross-taxonomy comparison. When a pipeline produces fewer NLs than there are references, collisions are unavoidable: $\mathrm{Collapse}\ge\max(0,m-n)$. Configurations with $n<m$ therefore start from a nonzero floor and are not comparable to those with $n\ge m$.

The metrics correlate only in part, capture distinct failure modes and respond differently to the number of generated labels, so we retain all four (full correlation analysis in App.~\ref{sec:metrics-discussion}).

  \begin{table*}[ht]
  \centering
  \small
  \setlength{\tabcolsep}{4pt}
  \resizebox{\textwidth}{!}{%
  \begin{tabular}{llrlrlrr}
  \toprule
  \textbf{ID} & \textbf{Dataset} & \textbf{$N$} & \textbf{Genre} & \textbf{$\diameter$W} & \textbf{Lang.} & \textbf{Dis/Neut} & \textbf{\#
  Narr.} \\
  \midrule
  CA & CARDS \cite{Coan2021}             & 28{,}945 & Blogs / think-tank articles    &  50 & English             &  31 / 69 &  27 \\
  CO & Climate Obstruction \cite{2024co} &  1{,}330 & Social media ads       &  29 & English             &  74 / 26 &   7 \\
  CV & COVID Conspiracy \cite{2024cv}    &  1{,}099 & Telegram messages      & 152 & German              &  47 / 53 &  14 \\
  EU & EU DisinfoTest \cite{2024eu}      &  1{,}344 & Disinfo subnarratives  &  30 & English             & 100 / 0  & 170 \\
  HP & HALT-PROP \cite{2025hp}           &  1{,}000 & News articles          & 438 & Lithuanian          &  84 / 16 &  11 \\
  NM & Narr. Media Framing \cite{otmakhova-frermann-2025-narrative} & 100 & News articles & 703 & English& 100 / 0 & 17 \\
  PN & PolyNarrative \cite{2025pn}       &  1{,}892 & News articles          & 403 & BG, EN, PT, HI, RU  &  81 / 19 &  74 \\
  UK & UKElectionNarratives \cite{2025uk}&      974 & Social media posts     &  40 & English             & 100 / 0  &  32 \\
  
  \bottomrule
  \end{tabular}}%
   \caption{Overview of disinformation narrative datasets. \textbf{W} = mean words per text. \textbf{Dis/Neut} = share of texts labelled with $\geq$1 disinformation narrative   vs.\ texts with no narrative (\%). \textbf{\# Narr.} = codebook taxonomy size. On PN, we use the SemEval 2025 release \citep{2025semeval}. The narrative count on PN excludes 14 ``Other'' labels from the original 88 labels set. EU DisinfoTest consists of narrative statements: 754 disinformation narratives and 590 credible narratives (44\%). Every statement is linked to one of the broad narratives that form the taxonomy, which we use as narrative labels.}
  \label{tab:datasets}

  \vspace{0.3em}
  \end{table*}

\subsubsection{Open-world validation}
\label{sec:openworld}

The third tier, \textit{narrative discovery}, relies on human validation due to the absence of reference data. Annotators are provided with guidelines (see App.~\ref{app:annotation-guidelines}) to annotate the following columns: (1) \textit{is\_disinfo\_narrative}, with the answers yes, no and unclear (unclear was allowed when an annotator could not decide after rereading; we count it as not-a-yes throughout). (2) \textit{label\_match}, a choice among six options: (2a) four taxonomy labels, namely the two labels most similar to the NL (by embedding similarity) from the corpus's own taxonomy and the two most similar from the reference taxonomy, (2b) \textit{Other} for NLs that lie within the topics the dataset covers (in-domain) but match none of the four, and (2c) \textit{None} for NLs outside those topics (out-of-domain). (3) A \textit{confidence} score ranging from 1 to 5.

For 2(a), we use Qwen3-Embedding-4B \cite{zhang2025qwen3embedding} to retrieve the top-2 candidates per NL from the own and reference taxonomies. This retrieval is meant to be an annotator aid: it bounds the comparison set to what annotators can plausibly handle (the EU taxonomy alone has 170 labels). 2(b), \emph{Other (in-domain)}, marks NLs that fit the datasets' topics but match none of the four retrieved labels, candidates for extending the existing taxonomies. 2(c), \emph{None (out-of-domain)}, marks NLs outside the datasets' topics, candidates for narrative discovery.

\paragraph{yes-rate ($\uparrow$).}
Fraction of NLs annotated as recognizable disinformation narratives ($\texttt{is\_disinfo\_narrative}\!=\!\text{yes}$).
 
\paragraph{match-rate.}
Diagnostic only. Reported so that
$\mathrm{yes}\!\approx\!\mathrm{match}+\mathrm{other}+\mathrm{none}$ holds. A reference the top-2 retrieval did not surface cannot be matched, so the absolute level reflects the retrieval and not the pipeline.

\begin{figure*}[t]
    \centering
    \includegraphics[width=\textwidth]{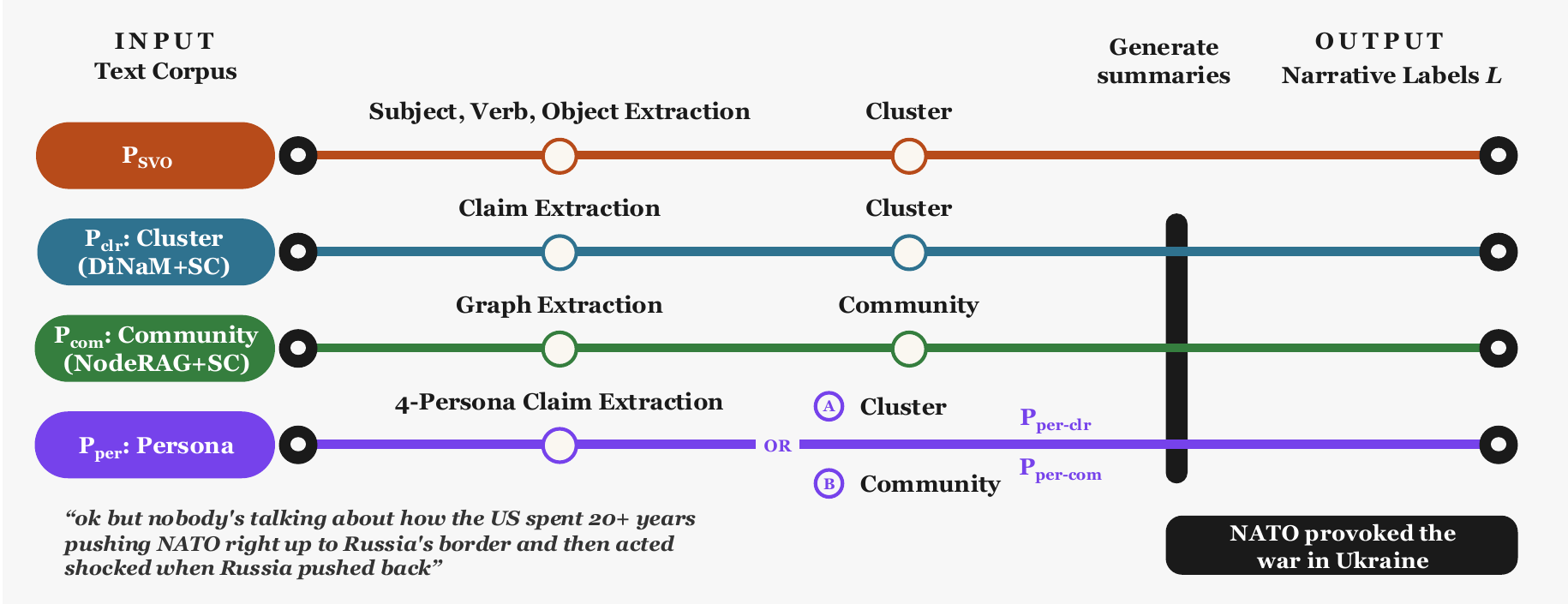}
    \caption{The pipeline configurations we compare (§\ref{sec:pipelines}). P$_{\text{per}}$ branches after extraction into P$_{\text{per-clr}}$ (\textbf{A}, clustering) and P$_{\text{per-com}}$ (\textbf{B}, graph + community detection). \emph{+SC} marks pipelines whose label-generation prompt we harmonized to the superclaim prompt of Appendix~\ref{sec:prompt}.}
    \label{fig:methodologies}
\end{figure*}

\paragraph{other-rate ($\uparrow$).} 
NLs in-domain but not covered by either taxonomy. It is an upper bound on the novel-narrative rate, since an NL marked \textit{Other} may have a match that dense retrieval did not return in the top-2. That bound depends on the embedding model and the taxonomy, not on the pipeline producing the NL, so we read \textit{other-rate} as a relative signal for comparing pipelines.
 
\paragraph{none-rate ($\uparrow$).}
NLs recognized as disinformation narratives but outside the datasets' domains. 

Although we work with multilingual datasets, all final labels are generated in English, the language of all referenced taxonomies. Two annotators with NLP backgrounds, one of them also with a humanities background, validated the outputs. This study is a pilot: it tests whether the discovery tier can identify valid narrative candidates, not how common they are, and it produces no gold labels. We therefore report agreement at two levels. At the strict level both annotators chose the label, and we treat these results as confident findings. At the union level at least one annotator did, and we keep these as lower-confidence evidence. A larger, independent panel that includes domain experts is needed before any candidate enters a taxonomy (see Limitations). Pairwise Cohen's $\kappa$ is reported in §\ref{sec:results_all}. Full per-cell numbers in Tab.~\ref{tab:iaa-all} (App.~\ref{app:iaa}).

The annotation sheet withheld all corpus-frequency information by design: annotators saw only the narrative label under judgment and the response fields, with no corpus-support counts and no community sizes. Displaying that information could have anchored judgments on frequency instead of on the label itself, so judgments at this tier cannot draw on within-corpus repetition evidence.

\subsection{Datasets}

Table~\ref{tab:datasets} presents an overview of the datasets we use for different evaluation settings. The datasets are used for evaluation on automated metrics. The pipeline results on PN and CO are also evaluated for narrative discovery. 

We use the multilingual corpora in their original languages and do not translate the source texts: documents are passed as they are into the first LLM call, whose prompt instructs the multilingual model (Gemma-4-31B-it) to produce English output (P$_{\text{com}}$ carries no explicit language instruction, but we verified that its outputs are English). Every following step therefore operates on English inputs, matching the language of all taxonomies we compare against, and the annotators of §\ref{sec:openworld} see English NLs only.

\subsection{Example Pipelines}
\label{sec:pipelines}

Figure~\ref{fig:methodologies} presents the pipeline families we compare as examples. All produce NLs from raw text but differ in (i) what is extracted from each document and (ii) how the extracted units are grouped before LLM summarization. We harmonized configurations like prompts or embedding models across the pipelines where components from existing systems were re-purposed (details in App.~\ref{sec:harmonize}).
 
\paragraph{P$_{\text{SVO}}$: SRL/Relatio-inspired no-LLM baseline.} Inspired by \citet{Ash_Gauthier_Widmer_2024} but using spaCy (\texttt{en\_core\_web\_lg} model) dependency parsing to extract Subject-Verb-Object (SVO) triplets in place of full SRL, and clustering whole triples instead of their arguments separately. The triples are embedded, UMAP-reduced and HDBSCAN-clustered. Each cluster's NL is the triplet closest to the cluster's mean embedding, making this a no-LLM baseline. \texttt{en\_core\_web\_lg} is an English parser, so P$_{\text{SVO}}$ is the one pipeline whose extraction step does not handle the multilingual corpora natively.
 
\paragraph{P$_{\text{clr}}$: clustering-centered, based on modified DiNaM.}
Following \citet{2025dinam}, an LLM extracts claims per document. The claims are embedded, UMAP-reduced and HDBSCAN-clustered, and an LLM generates a per-cluster NL.
 
\paragraph{P$_{\text{com}}$: graph + community.} 
We adapt the heterogeneous-graph construction of NodeRAG \citep{xu2025noderagstructuringgraphbasedrag}: an LLM decomposes each text into semantic units (proposition-level summaries of a text segment, in NodeRAG's terminology), entities, and relationships between them. A second LLM step generates attributes, descriptions attached to the structurally most important entities. Leiden \citep{Traag2019} community detection is run over the resulting graph. Per-community NLs are generated from the union of the semantic-unit and attribute texts of a community, which is the node selection NodeRAG uses for its own community summaries. Following NodeRAG, entities and relationships carry the graph structure but are not passed to the LLM at the summarization step. We use only NodeRAG's graph construction, not its retrieval layer.

\begin{table*}[t]
\centering
\small
\begin{tabular}{ll c rrrr rrrr}
\toprule
 & &  & \multicolumn{4}{c}{Narrative Recovery} & \multicolumn{4}{c}{Narrative Mining} \\
\cmidrule(lr){4-7} \cmidrule(lr){8-11}
Datasets & Method & \# NLs & Hung.$\uparrow$ & WCD$\downarrow$ & Coll.$\downarrow$ & C/R$\downarrow$
& Hung.$\uparrow$ & WCD$\downarrow$ & Coll.$\downarrow$ & C/R$\downarrow$
 \\
\midrule
CO$\rightarrow$ CARDS2  & P$_{\text{SVO}}$ & 27 & 0.586 & 0.468 & 2 & 0.286 & 0.462 & 0.497 & 29 & 0.906 \\
$R_{\text{own}}{=}7$  & P$_{\text{clr}}$ & 53 & 0.603 & 0.471 & 2 & 0.286 & 0.460 & 0.517 & \textbf{18} & \textbf{0.562} \\
$R_{\text{ref}}{=}32$ & P$_{\text{com}}$ & 267 & 0.632 & 0.492 & 2 & 0.286 & 0.515 & 0.512 & 21 & 0.656 \\
 & P$_{\text{per-clr}}$ & 186 & \textbf{0.649} & \textbf{0.447} & \textbf{0} & \textbf{0.000} & 0.517 & \textbf{0.477} & 23 & 0.719 \\
 & P$_{\text{per-com}}$ & 232 & 0.636 & 0.466 & \textbf{0} & \textbf{0.000} & \textbf{0.534} & 0.480 & 27 & 0.844 \\
\addlinespace
PN$\rightarrow$EU & P$_{\text{SVO}}$ & 182 & 0.553 & 0.464 & 45 & 0.608 & 0.495 & 0.481 & 123 & 0.724 \\
$R_{\text{own}}{=}74$ & P$_{\text{clr}}$ & 141 & 0.516 & 0.463 & 45 & 0.608 & \textbf{0.487} & 0.476 & 130 & 0.765 \\
$R_{\text{ref}}{=}170$& P$_{\text{com}}$ & 824 & 0.576 & 0.495 & 22 & 0.297 & 0.536 & 0.496 & \textbf{90} & \textbf{0.529} \\
 & P$_{\text{per-clr}}$ & 231 & 0.546 & \textbf{0.449} & 40 & 0.541 & 0.504 & \textbf{0.464} & 109 & 0.641 \\
 & P$_{\text{per-com}}$ & 392 & \textbf{0.577} & 0.462 & \textbf{19} & \textbf{0.257} & 0.535 & 0.466 & 97 & 0.571 \\

\bottomrule
\end{tabular}
\caption{Evaluation results. Single run per pipeline, on the two dataset pairs used for human validation. $X\rightarrow Y$ denotes corpus $X$ scored against reference taxonomy $Y$ in the mining setting. EU = EU DisinfoTest. CARDS2 = the revised CARDS taxonomy \citep{Coan2026} at its top two levels, used as a second, domain-specific reference for CO. $R$ is the number of reference labels, in the own or the reference taxonomy.}
\label{tab:selected_pipelines}
\end{table*}

\begin{table*}[t]
\centering\small
\setlength{\tabcolsep}{6pt}
\begin{tabular}{ll cccc c}
\toprule
Datasets & Method & \textbf{yes}$\uparrow$ & \textbf{match} & \textbf{other}$\uparrow$ & \textbf{none}$\uparrow$ & $2N$ \\
\midrule
CO$\rightarrow$CARDS2 & P$_{\text{per-clr}}$ & 62.10 (42.47) & 51.61 (38.17) & 10.48 (4.30)  & 0.00 (0.00) & 372 \\
                      & P$_{\text{per-com}}$ & 54.96 (37.07) & 36.64 (25.43) & 16.38 (10.34) & 1.94 (1.29) & 464 \\
\addlinespace[2pt]
PN$\rightarrow$EU     & P$_{\text{per-clr}}$ & 59.52 (42.64) & 28.14 (21.43) & 22.51 (14.72) & 8.66 (6.49) & 462 \\
                      & P$_{\text{per-com}}$ & 47.32 (37.24) & 26.28 (21.43) & 12.50 (9.31)  & 8.55 (6.51) & 784 \\
\bottomrule
\end{tabular}
\caption{Narrative discovery: human-validation label rates per run, annotators pooled (§\ref{sec:results_all}). $2N$: number of annotation rows ($N$ NLs $\times$ 2 annotators). Each cell is the \% of all rows. In parentheses, the \% of all rows on which that answer was given with confidence $\ge 4$. Rates are defined in §\ref{sec:openworld}. One \texttt{yes} row of PN P$_{\text{per-clr}}$ has no \texttt{label\_match} answer, so its three answer rates sum to slightly less than its yes-rate. \textbf{match} is diagnostic only and therefore carries no direction arrow.}
\label{tab:iaa-main-abs}
\end{table*}

\paragraph{P$_{\text{per}}$: Persona.} 
We apply multi-persona LLM claim extraction: four personas (investigative journalist, political scientist, conspiracy researcher, fact-checker) each independently extract up to three claims per document, on the premise that different analytical lenses yield different argumentative content from the same text. Claims are deduplicated (exact string match after whitespace stripping, pooled across all four personas) and then fed into either (a) the P$_{\text{clr}}$ clustering step or (b) the P$_{\text{com}}$ graph + community step. In (b) the NodeRAG-inspired heterogeneous graph is constructed over the extracted claims instead of raw documents.

\section{Results}
\label{sec:results_all}

\paragraph{Narrative Recovery and Mining}
The CO$\rightarrow$CARDS2 and PN$\rightarrow$EU cells in Table~\ref{tab:selected_pipelines} are representative of the full seven-dataset benchmark in Table~\ref{tab:all_pipelines_collapse} in the Appendix. We make 14 cluster-versus-graph comparisons: (P$_{\text{clr}}$ vs. P$_{\text{com}}$ and P$_{\text{per-clr}}$ vs. P$_{\text{per-com}}$ on each of the seven datasets). Extraction is identical within each pair, so differences are attributable to the grouping step. In recovery, a consistent trade-off appears: clustering wins on WCD in 13/14 comparisons (only UK/Persona reverses, by 0.002), graph aggregation wins on Hungarian in 12/14 (CO and HP reverse within the Persona pair only), and graph matches or wins over clustering on recovery Collapse in 14/14 (10 strict wins, 4 ties).
On the \emph{mining} side (X$\rightarrow$EU) the same trade-off holds but less uniformly: clustering wins WCD in 10/14 (graph reverses on CO for the P$_{\text{clr}}$-P$_{\text{com}}$ pair, on NM for both pairs, and on UK/Persona) and graph wins Hungarian in 10/14. The graph advantage is clearest on mining-Collapse, where the graph variant is lower in 14/14 comparisons (12/14 when CO is instead evaluated against CARDS2, Table~\ref{tab:all_pipelines_collapse_selected}, where both CO pairs reverse). The $P_{\text{SVO}}$ baseline never beats the graph pipelines on recovery Hungarian or on Collapse, yet it has the best mining Hungarian on CO and NM and beats some LLM pipelines on WCD. A no-LLM baseline outscoring LLM pipelines points to a limitation of the metrics (App.~\ref{sec:metrics-discussion}).

\paragraph{Narrative Discovery}
\label{sec:discovery-results}

We choose our two P$_{\text{per}}$ variants for human validation since they share most components and differ only in the aggregation step, isolating the clustering-vs-graph contribution. Both yield close results on automated metrics (P$_{\text{per-clr}}$ vs. P$_{\text{per-com}}$: $\Delta\mathrm{Hun}\le0.03$, $\Delta\mathrm{WCD}\le0.02$ in Table~\ref{tab:selected_pipelines}), while producing fewer NLs, making them a suitable choice for human validation.

Both annotators labeled every candidate independently, and Tab.~\ref{tab:iaa-main-abs} pools their annotations per run (\emph{merged}): $N$ candidates yield $2N$ annotation rows, and rates are computed over these rows. On this basis the table reports annotation rates: \texttt{yes} 47--62\%, \texttt{other} 10--23\%, \texttt{none} 0--9\%. Agreement is comparable to other narrative-related annotations, e.g., PolyNarrative reports Krippendorff's $\alpha{=}0.571$ (coarse) / $0.480$ (fine), with our $\kappa_{\texttt{merged}}$ in 0.56--0.65 and $\kappa_{\texttt{label\_match}}$ in 0.39--0.52. Here $\kappa_{\texttt{merged}}$ is Cohen's $\kappa$ on the collapsed three-category space \{in-domain (candidate match or \texttt{Other}), \texttt{None}, not-a-yes\}, and $\kappa_{\texttt{label\_match}}$ is Cohen's $\kappa$ on the raw \texttt{label\_match} answers. The comparison is approximate due to setup differences (full per-cell numbers and discussion in App.~\ref{app:iaa}).

Aggregate \texttt{none} rates are similar within each dataset across the two persona pipelines (CO: $0.00\%$ vs.\ $1.94\%$; PN: $8.66\%$ vs.\ $8.55\%$, Tab.~\ref{tab:iaa-main-abs}). The aggregate rates, however, mask very large disagreement on which candidates are flagged: Writing $a$ and $b$ for the sets of NLs each annotator marked \texttt{None}, the overlap $|a \cap b| / |a \cup b|$ is $9/31$ and $14/53$ on PN P$_{\text{per-clr}}$ / P$_{\text{per-com}}$, and $1/8$ on CO P$_{\text{per-com}}$. The two annotators thus validate largely disjoint sets of NLs as out-of-domain. We discuss a possible explanation of this divergence in §\ref{sec:discovered}.

In the LLM-judge experiments (App.~\ref{sec:llm-judge}), the two LLM judges agree more with each other than either agrees with the humans. As deployed here, they cannot replace human validation at the discovery tier, although their unanimous votes might be used as a filter.

\section{Analysis}
The two pipeline families yield close results on automated metrics. This section examines what separates them beyond those. Correlations between metrics are discussed in App.~\ref{sec:metrics-discussion}.

\subsection{Singletons and the repeatability assumption}
\label{sec:singleton}
\begin{table}[h]
\centering
\small
\begin{tabular}{lrr}
\toprule
agreement slice & CO & PN \\
\midrule
both: \emph{None (out-of-domain)}        &  0 / 1  &  5 / 14 \\
either: \emph{None}                      &  4 / 8  & 20 / 53 \\
both: \emph{Other (in-domain, no match)} &  2 / 14 &  6 / 22 \\
either: \emph{Other}                     & 16 / 62 & 24 / 76 \\
\bottomrule
\end{tabular}
\caption{Singletons among the NLs of P$_{\text{per-com}}$ in each annotator-agreement group. Each cell gives singletons / all NLs in the group. \emph{both}: both annotators gave the label. \emph{either}: at least one did.}
\label{tab:singleton-counts}
\end{table}

Under the criterion of §\ref{sec:relatedwork}, a singleton still qualifies as a narrative label instead of a single claim instance when its core message could subsume further claim instances, even if only one of them occurs in the corpus at hand.
Singletons are common in graph output (70/232 = 30\% on CO; 149/392 = 38\% on PN) and persist across every annotator-agreement slice with more than one item (Tab.~\ref{tab:singleton-counts}; candidates grouped by the strict and union levels of §\ref{sec:openworld}, i.e. both annotators agreeing vs. either). Even under our strictest filter, with both annotators agreeing on \emph{Other (in-domain, no match)}, 14\% of CO and 27\% of PN graph-pipeline validated narrative labels are singletons. The clustering pipelines cannot reproduce singletons. Lowering HDBSCAN's \texttt{min\_cluster\_size} from 25 (default value in DiNaM) does not reproduce them, because the structural analog of a graph singleton is HDBSCAN's noise pile (unclustered single-instance claims dropped before summarization), not a size-2 cluster. The latter still requires two semantically similar claims, whereas a singleton arises from a single isolated one. Reducing \texttt{min\_cluster\_size} to 2 produces 4069 narrative labels on PN (Tab.~\ref{tab:sweep-pn}), exceeding the practical capacity of human validation.

This has two consequences. (1) The repeatability assumption shifts its reference between mining and discovery. In mining, a pipeline counts repetition inside the corpus. Our annotators saw no corpus-frequency counts (§\ref{sec:openworld}), so their acceptance of these labels cannot rest on within-corpus repetition. Our reading of what it rests on instead is prior exposure to similar narratives in news and broader disinformation discourse. Alternatives we cannot rule out are that a label reads as plausible on its own, that acceptance reflects the guidelines' example-based grounding (App.~\ref{app:annotation-guidelines}), or that the absence of occurrence information itself invites acceptance. Under any of these, the judgment draws on something other than within-corpus repetition. (2) Narrative mining pipelines conditioned on closed-corpus repetition can systematically miss the long tail of novel narrative candidates whose corpus evidence is sparse, the same long tail that humans identify by drawing on open-world knowledge.

Additional experiments show that the singleton finding is robust across seeds, pipelines, and persona settings. The related experiments are documented in App.~\ref{sec:robustness-singletons}.

\subsection{Parameter sensitivity and topical coverage}
\label{sec:param-sweep}

\begin{table}[t] 
\centering
\small 
\begin{tabular}{lrrrrr}
\toprule 
Setting & $N$ & ukr\% & cli\% & H$\uparrow$ & Coll$\downarrow$ \\
\midrule 
\multicolumn{6}{l}{\textit{HDBSCAN (mcs, ms)}} \\
(2, 1)& 4069 & 55.8 & 27.3 & 0.690 & 13 \\ 
(5, 2)& 1300 & 52.1 & 29.5 & 0.673 & 25 \\
(15, 10)&332 & 52.1 & 33.7 & 0.645 & 34 \\ 
(25, 20)$^{*}$&141 & 83.0 &2.1 & 0.582 & 51 \\ 
\midrule 
\multicolumn{6}{l}{\textit{Leiden modularity-$\gamma$}} \\ 
$\gamma\!=\!0.1$&558 & 19.7 & 36.0 & 0.631 & 44 \\ 
$\gamma\!=\!1^{*}$&824 & 38.1 & 28.8 & 0.651 & 30 \\ 
$\gamma\!=\!5$& 1362 & 45.7 & 31.3 & 0.668 & 24 \\ 
$\gamma\!=\!10$ & 1938 & 49.9 & 29.8 & 0.671 & 26 \\ 
\bottomrule
\end{tabular}
\caption{Parameter sweep digest: label count vs.\ topical balance vs.\ recovery on PN (full sweep in Table~\ref{tab:sweep-pn}). HDBSCAN runs on P$_{\text{clr}}$ and Leiden modularity-$\gamma$ on P$_{\text{com}}$. (mcs, ms): HDBSCAN's \texttt{min\_cluster\_size} and \texttt{min\_samples}. $^{*}$=algorithm default. ukr\%/cli\%: share of NLs in the War in Ukraine / climate topics. H: Hungarian similarity (recovery, $\uparrow$). Coll: Collapse (recovery, $\downarrow$, out of $m{=}74$).  
}
\label{tab:pn_sweep_slim} 
\end{table}

A topical coverage issue became visible during human validation on PN: the persona-cluster pipeline produced labels almost entirely about Russia's war in Ukraine, with climate-denial narratives as the second main topic of the dataset nearly absent from the output.  Table~\ref{tab:pn_sweep_slim} traces the parameter regime that produces the mismatch (full sweep in App. Table~\ref{tab:sweep-pn}, the LLM-based topical coverage procedure is presented in App.~\ref{app:topic}).

We sweep each of the four aggregation pipelines independently over the granularity parameter of its own algorithm, two clustering-based (P$_{\text{clr}}$, P$_{\text{per-clr}}$ over HDBSCAN's \texttt{min\_cluster\_size}) and two graph-based (P$_{\text{com}}$, P$_{\text{per-com}}$ over Leiden's modularity $\gamma$). Both clustering pipelines suppress the minority topic at their algorithm default. At DiNaM's default \texttt{min\_cluster\_size}=25, the climate-topic share of generated labels on P$_{\text{clr}}$ collapses to 2.1\%, versus 27--34\% at lower granularities, and P$_{\text{per-clr}}$ collapses to 1.7\%, versus 25--27\% (App.~Tab.~\ref{tab:sweep-pn}). The cause is a thresholding artifact: HDBSCAN with a high minimum cluster size treats minority-topic groups as noise and drops them before summarization, leaving only the dominant topic in the output. Lowering \texttt{min\_cluster\_size} restores topical balance at the cost of more labels (332 at mcs=15 and 4069 at mcs=2 on P$_{\text{clr}}$, App.~Tab.~\ref{tab:sweep-pn}). The default that prior work adopts is the setting that loses the minority topic.

Neither graph pipeline exhibits this collapse. At Leiden's algorithm-default $\gamma=1$, P$_{\text{com}}$ produces a 38\%/29\% Ukraine/climate split and P$_{\text{per-com}}$ a 37\%/38\% split. Across the full $\gamma$-sweep ($0.1$ to $10$) neither topic drops below 19.7\% on either pipeline, so the split shifts with granularity but no topic is suppressed at any setting we tried (App.~Tab.~\ref{tab:sweep-pn}). The graph approach is therefore more resilient to its own default parameters with respect to topical coverage on this corpus.

This asymmetry is conditional on the specific algorithms and on the corpus topic distribution: in a single-topic corpus the HDBSCAN behavior would not register as a problem. The practical observation is that the default parameters of these approaches have very different consequences for topical coverage on multi-topic corpora, and that this consequence is invisible to the automated recovery and mining metrics.

\subsection{Discovered Narrative Candidates}
\label{sec:discovered}

App.~\ref{sec:app-discovered} documents which concrete narrative candidates were identified by human validation. On CO, both annotators selected \textit{None} (out-of-domain) on 1 item, and at least one annotator selected \textit{None} on 8 items. Both annotators selected \textit{Other} (in-domain, no match) on 16 items. Taking the union of the \textit{Other}- and \textit{None}-flagged items, a repeated pattern emerges: many candidates share a core message, arguing against state regulation and for letting the market decide sustainability measures. In total the CO human validation yielded 24 candidates (16 both-\textit{Other} and 8 either-\textit{None}, disjoint sets), of which 6 group together at one level of abstraction within the CO taxonomy. 

The largest of these, ``\textit{Market-driven energy production is superior to government regulation and restrictive environmental mandates}'' (99 source ads, Tab.~\ref{tab:co-other}), was marked \textit{Other} by both annotators: it lies in the climate-obstruction domain but matched none of the retrieved CO or CARDS2 labels, and no CO entry addresses regulation or markets.

We propose a new narrative category candidate for this group, drawing on the phenomenon as discussed in works such as \citet{Jacques01062008}: \textit{State Regulation \& Free Market: Argues against state regulation methods and potential pitfalls and/or for free market forces being more suitable for handling energy transition processes}.

On PN, the candidates currently work better as a flat list of disinformation narrative labels than as a coherent group with a clear umbrella category, and the scope question (in- vs.\ out-of-domain relative to the PN taxonomy) is less clear than on CO. Both annotators marked ``\textit{National security depends on total self-reliance in energy and military production to eliminate foreign strategic vulnerabilities}'' (59 source documents, Tab.~\ref{tab:pn-ood}) as \textit{None}, although its self-reliance framing is close to the war-in-Ukraine discourse PN covers, a scope judgment that calls for expert assessment at scale. On these two corpora the contrast is: a focused single-domain corpus with a small taxonomy (CO: 7 labels, climate obstruction only) yields a coherent extension candidate, while a broader multi-topic corpus with a larger taxonomy (PN: 74 labels across multiple topics and languages) yields extension candidates but no single coherent group.

The per-annotator overlap pattern from §\ref{sec:discovery-results} generalizes across the label space. Set overlap $|a \cap b|/|a \cup b|$ on \texttt{yes} holds in the 0.68--0.74 band across all four runs but drops to 0.05--0.41 on \texttt{other} and 0.13--0.29 on \texttt{none} (App.~Tab.~\ref{tab:iaa-all}). We read this in line with the interpretation discussed in §\ref{sec:singleton}: a \texttt{yes} paired with a retrieved candidate label gives both annotators a common reference point inside the taxonomies. For \texttt{other} and \texttt{none} there is no such anchor, and each annotator falls back on what they have encountered outside the corpus. If that is what drives singleton acceptance, it would also explain why the two annotators diverge on narratives outside the taxonomies.

In the recent CARDS2 revision \citep{Coan2026}, not only new categories were added but also existing ones re-grouped and revised. A similar process is imaginable here, where the semi-automatically identified candidates could serve as a starting point.

\section{Conclusions}
\label{sec:conclusions}

We return to our three research questions.

\textbf{RQ1.} Different automated metrics emphasize different strengths of the two pipeline families we compared, but further analysis indicated two structural differences that closed-world metrics do not capture. At default parameters, the cluster-based pipelines left one of PolyNarrative's two main topics severely underrepresented, whereas the graph-based pipelines kept both topics across the parameter sweep. The graph-based pipelines also produced singletons at substantial rates (30--62\% of narrative labels across our main configurations), which cluster-based pipelines cannot reproduce. 

\textbf{RQ2.} Many graph-pipeline singletons are validated as disinformation narrative candidates. This questions whether the repeatability assumption of narrative mining, that a narrative is recognized by within-corpus repetition, extends to discovery. Our interpretation is that it does, but with a shift in what the repetition refers to. Annotators appear to recognize these narratives from the broader discourse and not from the corpus at hand. Regardless of that interpretation, annotators accepted such labels without any within-corpus frequency evidence.

\textbf{RQ3.} Narrative-mining pipelines can surface novel candidates for narrative-label extensions. We release these sets for CO and PN to support taxonomy maintenance.

\section*{Limitations}

\textbf{Pilot-scale annotation and annotator bias.} The discovery tier relies on two annotators with NLP backgrounds, one of them also with a humanities background, applied to two datasets (CO, PN). The pilot shows that the tier identifies narrative candidates in pipeline output, but not how much agreement should be required before a candidate is added to a taxonomy. Both annotators are also co-authors of this paper, which carries two risks the pilot cannot rule out. Familiarity with the project may bias them toward accepting pipeline outputs (confirmation bias). Withholding frequency information (§\ref{sec:openworld}) and grounding each judgment in retrieved labels from expert-curated taxonomies constrain this without removing it. The NLP background they share may also limit which candidates they recognize as narratives at all, which directly affects the singleton reading of §\ref{sec:singleton}. An interdisciplinary annotator panel is therefore the necessary next step, since no single domain expertise covers climate obstruction, the war in Ukraine and COVID-19 alike, and the results in App.~\ref{sec:llm-judge} indicate that LLM judges we tested are not a viable substitute under current capabilities.

\textbf{Topical coverage.} Our findings point to limitations of clustering configurations in regard to balanced topical coverage. This finding is based on one dataset with two main topics (PN) and remains to be explored on further multi-topic datasets. 

\textbf{Semantic equivalence.} The closed-world metrics rely on embedding similarity for matching narrative labels to reference labels. \citet{sun2025textembeddingscaptureimplicit} note that current embedding models capture surface meaning more reliably than implicit semantics. We mitigate this with multiple complementary metrics (§\ref{sec:metrics}) but cannot fully decouple metric outcomes from the embedding model's behavior.

\textbf{Temporal bias and knowledge leakage.} Several of our corpora cover topics that are extensively represented in recent LLM training data. COVID-19 is the sharpest case: unlike climate change it is tied to a specific historical event, so a model trained on post-2020 text may already encode the narratives and their later development, and can appear to anticipate an ``emerging'' narrative through hindsight instead of inference from the corpus. The same concern applies to climate-related corpora and to the war in Ukraine. Our design does not mitigate this leakage but holds it constant: every LLM-based configuration uses the same backbone, so leakage shared by all configurations is unlikely to explain the between-family differences our central findings rest on (topical coverage, singleton rates), which arise in the grouping steps and not in an LLM call. It does, however, limit what any absolute novelty claim can mean, which is why we phrase novelty relative to the reference taxonomy throughout. A stronger test would evaluate models whose training data predates the events in question. The underlying issue is also not only methodological: disinformation narratives do not emerge in a vacuum but within a socio-historical context and typically build on earlier narratives (COVID-19 narratives, for instance, continue a long anti-vaccination lineage), so it is an open question when a narrative should count as new and when as an updated version of an existing one.

\textbf{Prompt strategy.} We harmonized the label-generation prompt across all LLM-based pipelines to attribute differences to the grouping step instead of to prompt wording, but did not explore prompt optimization techniques. Prompt design affects narrative-label quality, especially on multilingual input: App.~\ref{sec:harmonize} shows measurable Collapse differences between our superclaim prompt and DiNaM's, and App.~\ref{sec:robustness-singletons} quantifies how much singleton counts vary between repeated LLM runs. A systematic study of prompting strategies for narrative-label generation is left to future work.

\section*{Ethics Statement}

All experiments use publicly released disinformation narrative datasets under their original licenses. We do not redistribute the original raw documents and only release their ids, which can be used to reconstruct the traced originals. The datasets and thus our generated narrative labels contain references to public figures (like politicians) and institutions in the context of disinformation discourse, but not private individuals, identifiers, or contact information. 

We release human-validated narrative-label candidates for the Climate Obstruction and PolyNarrative datasets, as discussion material for taxonomy maintenance and extension. The released narrative candidates are explicitly not gold labels, and not intended as training data for systems that generate or amplify disinformation framings. The discovery-tier human-validation pilot was carried out by two members of the research team, who are co-authors of this paper and were informed in advance of the nature of the content. Annotation was conducted as part of regular research activity, with no separate recruitment or compensation.

Claude Opus 4.6, 4.7 and 5 (including via Claude Code) were used for editing and reformulation of the manuscript and as a coding assistant during pipeline development and analysis. The authors take full responsibility for the final content.

\section*{Acknowledgments}

The work on this paper is performed in the scope of the projects ``VeraXtract'' (16IS24066), funded by the German Federal Ministry for Research, Technology and Aeronautics (BMFTR) and ``FIMI RESIST'' (reference: HORIZON-CL2-2025-01-101285890), funded by the European Union.

\bibliography{custom}

\begin{thebibliography}{34}
\providecommand{\natexlab}[1]{#1}

\bibitem[{Ash et~al.(2024)Ash, Gauthier, and Widmer}]{Ash_Gauthier_Widmer_2024}
Elliott Ash, Germain Gauthier, and Philine Widmer. 2024.
\newblock \href {https://doi.org/10.1017/pan.2023.8} {Relatio: Text semantics
  capture political and economic narratives}.
\newblock \emph{Political Analysis}, 32(1):115–132.

\bibitem[{Coan et~al.(2021)Coan, Boussalis, Cook, and Nanko}]{Coan2021}
Travis~G. Coan, Constantine Boussalis, John Cook, and Mirjam~O. Nanko. 2021.
\newblock \href {https://doi.org/10.1038/s41598-021-01714-4} {Computer-assisted
  classification of contrarian claims about climate change}.
\newblock \emph{Scientific Reports}, 11(1):22320.

\bibitem[{Coan et~al.(2026)Coan, Malla, Nanko, Kattrup, Roberts, Cook, and
  Boussalis}]{Coan2026}
Travis~G. Coan, Ranadheer Malla, Mirjam~O. Nanko, William Kattrup, J.~Timmons
  Roberts, John Cook, and Constantine Boussalis. 2026.
\newblock \href {https://doi.org/10.1038/s44458-025-00029-z} {Large language
  model reveals an increase in climate contrarian speech in the {United}
  {States} {Congress}}.
\newblock \emph{Communications Sustainability}, 1(1):37.

\bibitem[{Enevoldsen et~al.(2025)Enevoldsen, Chung, Kerboua, Kardos, Mathur,
  Stap, Gala, Siblini, Krzemiński, Winata, Sturua, Utpala, Ciancone,
  Schaeffer, Sequeira, Misra, Dhakal, Rystrøm, Solomatin, Ömer Çağatan,
  Kundu, Bernstorff, Xiao, Sukhlecha, Pahwa, Poświata, GV, Ashraf, Auras,
  Plüster, Harries, Magne, Mohr, Hendriksen, Zhu, Gisserot-Boukhlef, Aarsen,
  Kostkan, Wojtasik, Lee, Šuppa, Zhang, Rocca, Hamdy, Michail, Yang, Faysse,
  Vatolin, Thakur, Dey, Vasani, Chitale, Tedeschi, Tai, Snegirev, Günther,
  Xia, Shi, Lù, Clive, Krishnakumar, Maksimova, Wehrli, Tikhonova, Panchal,
  Abramov, Ostendorff, Liu, Clematide, Miranda, Fenogenova, Song, Safi, Li,
  Borghini, Cassano, Su, Lin, Yen, Hansen, Hooker, Xiao, Adlakha, Weller,
  Reddy, and Muennighoff}]{enevoldsen2025mmteb}
Kenneth Enevoldsen, Isaac Chung, Imene Kerboua, Márton Kardos, Ashwin Mathur,
  David Stap, Jay Gala, Wissam Siblini, Dominik Krzemiński, Genta~Indra
  Winata, Saba Sturua, Saiteja Utpala, Mathieu Ciancone, Marion Schaeffer,
  Gabriel Sequeira, Diganta Misra, Shreeya Dhakal, Jonathan Rystrøm, Roman
  Solomatin, and 67 others. 2025.
\newblock \href {https://arxiv.org/abs/2502.13595} {{MMTEB}: Massive
  multilingual text embedding benchmark}.
\newblock \emph{Preprint}, arXiv:2502.13595.

\bibitem[{Fei and Liu(2016)}]{fei2016breaking}
Geli Fei and Bing Liu. 2016.
\newblock \href {https://doi.org/10.18653/v1/N16-1061} {Breaking the closed
  world assumption in text classification}.
\newblock In \emph{Proceedings of the 2016 Conference of the North {A}merican
  Chapter of the Association for Computational Linguistics: Human Language
  Technologies}, pages 506--514, San Diego, California. Association for
  Computational Linguistics.

\bibitem[{{Gemma Team}(2026)}]{gemma_4_31b}
{Gemma Team}. 2026.
\newblock \href {https://arxiv.org/abs/2607.02770} {{Gemma 4} technical
  report}.
\newblock \emph{Preprint}, arXiv:2607.02770.
\newblock Model card: \url{https://huggingface.co/google/gemma-4-31B-it}.

\bibitem[{Gerard et~al.(2026)Gerard, Luceri, Blas, and Ferrara}]{Gerard2026}
Patrick Gerard, Luca Luceri, Leonardo Blas, and Emilio Ferrara. 2026.
\newblock Cross-platform narrative prediction: Leveraging platform-invariant
  discourse networks.
\newblock In \emph{Proceedings of the ACM Web Conference 2026}, WWW '26, page
  4898–4909, New York, NY, USA. Association for Computing Machinery.

\bibitem[{Grootendorst(2022)}]{2022bertopic}
Maarten Grootendorst. 2022.
\newblock \href {https://arxiv.org/abs/2203.05794} {{BERTopic}: Neural topic
  modeling with a class-based {TF-IDF} procedure}.
\newblock \emph{Preprint}, arXiv:2203.05794.

\bibitem[{Hanley et~al.(2024)Hanley, Kumar, and Durumeric}]{hanley2024specious}
Hans W.~A. Hanley, Deepak Kumar, and Zakir Durumeric. 2024.
\newblock Specious sites: Tracking the spread and sway of spurious news stories
  at scale.
\newblock In \emph{2024 IEEE Symposium on Security and Privacy (SP)}, pages
  1609--1627. IEEE.

\bibitem[{Haouari et~al.(2025)Haouari, Scarton, Faggiani, Nikolaidis, Kotseva,
  Abu~Farha, Linge, and Bontcheva}]{2025uk}
Fatima Haouari, Carolina Scarton, Nicolò Faggiani, Nikolaos Nikolaidis, Bonka
  Kotseva, Ibrahim Abu~Farha, Jens Linge, and Kalina Bontcheva. 2025.
\newblock \href {https://doi.org/10.1609/icwsm.v19i1.35950}
  {{UKElectionNarratives}: A dataset of misleading narratives surrounding
  recent {UK} general elections}.
\newblock \emph{Proceedings of the International AAAI Conference on Web and
  Social Media}, 19(1):2477--2495.

\bibitem[{Heinrich et~al.(2024)Heinrich, Blombach, Doan~Dang, Zilio,
  Havenstein, Dykes, Evert, and Sch{\"a}fer}]{2024cv}
Philipp Heinrich, Andreas Blombach, Bao~Minh Doan~Dang, Leonardo Zilio, Linda
  Havenstein, Nathan Dykes, Stephanie Evert, and Fabian Sch{\"a}fer. 2024.
\newblock \href {https://aclanthology.org/2024.lrec-main.173/} {Automatic
  identification of {COVID}-19-related conspiracy narratives in {G}erman
  {Telegram} channels and chats}.
\newblock In \emph{Proceedings of the 2024 Joint International Conference on
  Computational Linguistics, Language Resources and Evaluation (LREC-COLING
  2024)}, pages 1932--1943, Torino, Italia. ELRA and ICCL.

\bibitem[{Jacques et~al.(2008)Jacques, Dunlap, and Freeman}]{Jacques01062008}
Peter~J. Jacques, Riley~E. Dunlap, and Mark Freeman. 2008.
\newblock \href {https://doi.org/10.1080/09644010802055576} {The organisation
  of denial: Conservative think tanks and environmental scepticism}.
\newblock \emph{Environmental Politics}, 17(3):349--385.

\bibitem[{{Keith Norambuena} and Mitra(2021)}]{Norambuena2021NarrativeMaps}
Brian~Felipe {Keith Norambuena} and Tanushree Mitra. 2021.
\newblock \href {https://doi.org/10.1145/3432927} {Narrative maps: An
  algorithmic approach to represent and extract information narratives}.
\newblock \emph{Proceedings of the ACM on Human-Computer Interaction},
  4(CSCW3):1--33.

\bibitem[{Kuhn(1955)}]{kuhn1955hungarian}
Harold~W. Kuhn. 1955.
\newblock \href {https://doi.org/10.1002/nav.3800020109} {The {H}ungarian
  method for the assignment problem}.
\newblock \emph{Naval Research Logistics Quarterly}, 2(1--2):83--97.

\bibitem[{Luo(2005)}]{luo2005coreference}
Xiaoqiang Luo. 2005.
\newblock \href {https://aclanthology.org/H05-1004/} {On coreference resolution
  performance metrics}.
\newblock In \emph{Proceedings of Human Language Technology Conference and
  Conference on Empirical Methods in Natural Language Processing}, pages
  25--32, Vancouver, British Columbia, Canada. Association for Computational
  Linguistics.

\bibitem[{{Microsoft}(2026)}]{microsoft_harrier_v1_2026}
{Microsoft}. 2026.
\newblock {harrier-oss-v1-0.6b}: Multilingual text embedding model card.
\newblock Hugging Face model card,
  \url{https://huggingface.co/microsoft/harrier-oss-v1-0.6b}.
\newblock Released March 2026.

\bibitem[{Nikolaidis et~al.(2025)Nikolaidis, Stefanovitch, Silvano, Dimitrov,
  Yangarber, Guimar{\~a}es, Sartori, Androutsopoulos, Nakov, Da~San~Martino,
  and Piskorski}]{2025pn}
Nikolaos Nikolaidis, Nicolas Stefanovitch, Purifica{\c{c}}{\~a}o Silvano,
  Dimitar~Iliyanov Dimitrov, Roman Yangarber, Nuno Guimar{\~a}es, Elisa
  Sartori, Ion Androutsopoulos, Preslav Nakov, Giovanni Da~San~Martino, and
  Jakub Piskorski. 2025.
\newblock \href {https://doi.org/10.18653/v1/2025.acl-long.1513}
  {{P}oly{N}arrative: A multilingual, multilabel, multi-domain dataset for
  narrative extraction from news articles}.
\newblock In \emph{Proceedings of the 63rd Annual Meeting of the Association
  for Computational Linguistics (Volume 1: Long Papers)}, pages 31323--31345,
  Vienna, Austria. Association for Computational Linguistics.

\bibitem[{Otmakhova and Frermann(2025)}]{otmakhova-frermann-2025-narrative}
Yulia Otmakhova and Lea Frermann. 2025.
\newblock \href {https://doi.org/10.18653/v1/2025.findings-acl.477} {Narrative
  media framing in political discourse}.
\newblock In \emph{Findings of the Association for Computational Linguistics:
  ACL 2025}, pages 9167--9196, Vienna, Austria. Association for Computational
  Linguistics.

\bibitem[{Pham et~al.(2024)Pham, Hoyle, Sun, Resnik, and
  Iyyer}]{pham-etal-2024-topicgpt}
Chau~Minh Pham, Alexander Hoyle, Simeng Sun, Philip Resnik, and Mohit Iyyer.
  2024.
\newblock \href {https://doi.org/10.18653/v1/2024.naacl-long.164}
  {{T}opic{GPT}: A prompt-based topic modeling framework}.
\newblock In \emph{Proceedings of the 2024 Conference of the North American
  Chapter of the Association for Computational Linguistics: Human Language
  Technologies (Volume 1: Long Papers)}, pages 2956--2984, Mexico City, Mexico.
  Association for Computational Linguistics.

\bibitem[{Piper et~al.(2021)Piper, So, and Bamman}]{piper-etal-2021-narrative}
Andrew Piper, Richard~Jean So, and David Bamman. 2021.
\newblock \href {https://doi.org/10.18653/v1/2021.emnlp-main.26} {Narrative
  theory for computational narrative understanding}.
\newblock In \emph{Proceedings of the 2021 Conference on Empirical Methods in
  Natural Language Processing}, pages 298--311, Online and Punta Cana,
  Dominican Republic. Association for Computational Linguistics.

\bibitem[{Piskorski et~al.(2025)Piskorski, Mahmoud, Nikolaidis, Campos,
  Mario~Jorge, Dimitrov, Silvano, Yangarber, Sharma, Chakraborty, Guimaraes,
  Sartori, Stefanovitch, Xie, Nakov, and Da~San~Martino}]{2025semeval}
Jakub Piskorski, Tarek Mahmoud, Nikolaos Nikolaidis, Ricardo Campos, Alipio
  Mario~Jorge, Dimitar Dimitrov, Purifica{\c{c}}{\~a}o Silvano, Roman
  Yangarber, Shivam Sharma, Tanmoy Chakraborty, Nuno Guimaraes, Elisa Sartori,
  Nicolas Stefanovitch, Zhuohan Xie, Preslav Nakov, and Giovanni
  Da~San~Martino. 2025.
\newblock \href {https://aclanthology.org/2025.semeval-1.331/} {{S}em{E}val
  2025 task 10: Multilingual characterization and extraction of narratives from
  online news}.
\newblock In \emph{Proceedings of the 19th International Workshop on Semantic
  Evaluation (SemEval-2025)}, pages 2610--2643, Vienna, Austria. Association
  for Computational Linguistics.

\bibitem[{Pournaki and Willaert(2025)}]{pournaki-willaert-2025}
Armin Pournaki and Tom Willaert. 2025.
\newblock \href {https://doi.org/10.1057/s41599-025-06017-x} {Extracting
  narrative signals from public discourse: a network-based approach}.
\newblock \emph{Humanities and Social Sciences Communications}, 12(1):1774.

\bibitem[{{Qwen Team}(2026)}]{qwen2026qwen35}
{Qwen Team}. 2026.
\newblock \href {https://qwen.ai/blog?id=qwen3.5} {{Qwen3.5}: Towards native
  multimodal agents}.
\newblock Model card: \url{https://huggingface.co/Qwen/Qwen3.5-27B}.

\bibitem[{Rizgelienė et~al.(2026)Rizgelienė, Zubaitienė, Maliukevičius, and
  Marcinkevičius}]{2025hp}
Ieva Rizgelienė, Vilma Zubaitienė, Nerijus Maliukevičius, and Virginijus
  Marcinkevičius. 2026.
\newblock \href {https://doi.org/10.1038/s41597-025-06367-w} {{HALT-PROP}:
  Human-annotated {Lithuanian} textual corpus for propaganda narratives and
  techniques}.
\newblock \emph{Scientific Data}, 13(1):47.

\bibitem[{Rowlands et~al.(2024)Rowlands, Morio, Tanner, and Manning}]{2024co}
Harri Rowlands, Gaku Morio, Dylan Tanner, and Christopher Manning. 2024.
\newblock \href {https://doi.org/10.18653/v1/2024.findings-acl.330} {Predicting
  narratives of climate obstruction in social media advertising}.
\newblock In \emph{Findings of the Association for Computational Linguistics:
  ACL 2024}, pages 5547--5558, Bangkok, Thailand. Association for Computational
  Linguistics.

\bibitem[{Sosnowski et~al.(2024)Sosnowski, Modzelewski, Skorupska, Otterbacher,
  and Wierzbicki}]{2024eu}
Witold Sosnowski, Arkadiusz Modzelewski, Kinga Skorupska, Jahna Otterbacher,
  and Adam Wierzbicki. 2024.
\newblock \href {https://doi.org/10.18653/v1/2024.findings-emnlp.862} {{EU}
  {D}isinfo{T}est: a benchmark for evaluating language models' ability to
  detect disinformation narratives}.
\newblock In \emph{Findings of the Association for Computational Linguistics:
  EMNLP 2024}, pages 14702--14723, Miami, Florida, USA. Association for
  Computational Linguistics.

\bibitem[{Sosnowski et~al.(2025)Sosnowski, Modzelewski, Skorupska, and
  Wierzbicki}]{2025dinam}
Witold Sosnowski, Arkadiusz Modzelewski, Kinga Skorupska, and Adam Wierzbicki.
  2025.
\newblock \href {https://doi.org/10.18653/v1/2025.emnlp-main.1537}
  {{D}i{N}a{M}: Disinformation narrative mining with large language models}.
\newblock In \emph{Proceedings of the 2025 Conference on Empirical Methods in
  Natural Language Processing}, pages 30224--30251, Suzhou, China. Association
  for Computational Linguistics.

\bibitem[{Sun et~al.(2025)Sun, Huang, Tung, and
  Yu}]{sun2025textembeddingscaptureimplicit}
Yiqun Sun, Qiang Huang, Anthony K.~H. Tung, and Jun Yu. 2025.
\newblock \href {https://arxiv.org/abs/2506.08354} {Position: Text embeddings
  should capture implicit semantics, not just surface meaning}.
\newblock \emph{Preprint}, arXiv:2506.08354.

\bibitem[{Tangherlini et~al.(2020)Tangherlini, Shahsavari, Shahbazi,
  Ebrahimzadeh, and Roychowdhury}]{Tangherlini2020}
Timothy~R. Tangherlini, Shadi Shahsavari, Behnam Shahbazi, Ehsan Ebrahimzadeh,
  and Vwani Roychowdhury. 2020.
\newblock \href {https://doi.org/10.1371/journal.pone.0233879} {An automated
  pipeline for the discovery of conspiracy and conspiracy theory narrative
  frameworks: {Bridgegate}, {Pizzagate} and storytelling on the {Web}}.
\newblock \emph{PLOS ONE}, 15(6):e0233879.

\bibitem[{Traag et~al.(2019)Traag, Waltman, and van Eck}]{Traag2019}
V.~A. Traag, L.~Waltman, and N.~J. van Eck. 2019.
\newblock \href {https://doi.org/10.1038/s41598-019-41695-z} {From {L}ouvain to
  {L}eiden: guaranteeing well-connected communities}.
\newblock \emph{Scientific Reports}, 9(1):5233.

\bibitem[{Upravitelev et~al.(2026)Upravitelev, Solopova, Jakob, Sahitaj,
  M{\"o}ller, and Schmitt}]{upravitelev-etal-2026-retrieving}
Max Upravitelev, Veronika Solopova, Charlott Jakob, Premtim Sahitaj, Sebastian
  M{\"o}ller, and Vera Schmitt. 2026.
\newblock \href {https://doi.org/10.63317/5markfecdiyu} {Retrieving climate
  change disinformation by narrative}.
\newblock In \emph{Proceedings of the 2nd Workshop on Ecology, Environment, and
  Natural Language Processing}, pages 1--14, Palma de Mallorca, Spain. European
  Language Resources Association.

\bibitem[{Vaze et~al.(2022)Vaze, Han, Vedaldi, and
  Zisserman}]{vaze2022generalized}
Sagar Vaze, Kai Han, Andrea Vedaldi, and Andrew Zisserman. 2022.
\newblock Generalized category discovery.
\newblock In \emph{Proceedings of the IEEE/CVF Conference on Computer Vision
  and Pattern Recognition (CVPR)}, pages 7492--7501.

\bibitem[{Xu et~al.(2025)Xu, Zheng, Li, Chen, Liu, Chen, and
  Sun}]{xu2025noderagstructuringgraphbasedrag}
Tianyang Xu, Haojie Zheng, Chengze Li, Haoxiang Chen, Yixin Liu, Ruoxi Chen,
  and Lichao Sun. 2025.
\newblock \href {https://arxiv.org/abs/2504.11544} {{Node{RAG}}: Structuring
  graph-based {RAG} with heterogeneous nodes}.
\newblock \emph{Preprint}, arXiv:2504.11544.

\bibitem[{Zhang et~al.(2025)Zhang, Li, Long, Zhang, Lin, Yang, Xie, Yang, Liu,
  Lin, Huang, and Zhou}]{zhang2025qwen3embedding}
Yanzhao Zhang, Mingxin Li, Dingkun Long, Xin Zhang, Huan Lin, Baosong Yang,
  Pengjun Xie, An~Yang, Dayiheng Liu, Junyang Lin, Fei Huang, and Jingren Zhou.
  2025.
\newblock \href {https://arxiv.org/abs/2506.05176} {Qwen3 embedding: Advancing
  text embedding and reranking through foundation models}.
\newblock \emph{Preprint}, arXiv:2506.05176.

\end{thebibliography}

\appendix
\raggedbottom 

\clearpage
\section{Terminology}
\label{sec:terminology}
\begin{table}[h]
\centering\small
\renewcommand{\arraystretch}{1.25}
\begin{tabularx}{\columnwidth}{@{}l X@{}}
\toprule
\textbf{Term} & \textbf{Definition} \\
\midrule
Narrative label (NL) & All labels generated by a pipeline \\
Narrative candidate  & An NL that human validation marked as in-domain-but-unmatched (\emph{Other}) or out-of-domain (\emph{None}), i.e.\ a candidate for extending a taxonomy. \\
Singleton            & An NL whose underlying graph community contains a single node: one source claim that community detection could not merge with any other. \\
Reference label      & A label of the taxonomy an NL set is scored against ($R_{\text{own}}$ for the corpus's own taxonomy, $R_{\text{ref}}$ for the independent one). \\
Superclaim           & A normalized declarative assertion which a reader could agree or disagree with, and which expresses a message general enough to group multiple individual texts under it (§\ref{sec:relatedwork}). Our criterion for counting a generated output as a narrative label and not as a single claim instance. \\
\bottomrule
\end{tabularx}
\caption{Terminology used throughout the paper.}
\label{tab:terminology}
\end{table}

\clearpage
\section{Full Methods Table}
\label{sec:full-methods}

\begin{table*}[!t]
\centering
\small
\begin{tabular}{ll r rrrr rrrr}
\toprule
 & & \# Narrative & \multicolumn{4}{c}{Recovery} & \multicolumn{4}{c}{Mining} \\
\cmidrule(lr){4-7} \cmidrule(lr){8-11}
DS & Method & Labels & Hung.$\uparrow$ & WCD$\downarrow$ & Coll.$\downarrow$ & C/R$\downarrow$ & Hung.$\uparrow$ & WCD$\downarrow$ & Coll.$\downarrow$ & C/R$\downarrow$ \\
\midrule
CA$\rightarrow$EU & P$_{\text{SVO}}$ & 452 & 0.606 & 0.497 & 6 & 0.222 & 0.514 & 0.491 & 116 & 0.682 \\
R$_{\text{own}}$=27 & P$_{\text{dinam}}$ & 253 & 0.636 & \textbf{0.434} & 7 & 0.259 & 0.522 & \textbf{0.446} & 116 & 0.682 \\
R$_{\text{ref}}$=170 & P$_{\text{clr}}$ & 740 & 0.620 & 0.484 & 2 & 0.074 & 0.515 & 0.492 & 86 & 0.506 \\
 & P$_{\text{com}}$ & 3481 & 0.638 & 0.507 & \textbf{1} & \textbf{0.037} & 0.547 & 0.503 & 81 & 0.476 \\
 & P$_{\text{per-clr}}$ & 1622 & 0.631 & 0.476 & 2 & 0.074 & 0.543 & 0.476 & 80 & 0.471 \\
 & P$_{\text{per-com}}$ & 3259 & \textbf{0.641} & 0.483 & 2 & 0.074 & \textbf{0.555} & 0.483 & \textbf{65} & \textbf{0.382} \\
\addlinespace
CO$\rightarrow$EU & P$_{\text{SVO}}$ & 27 & 0.586 & 0.468 & 2 & 0.286 & \textbf{0.493} & 0.517 & 154 & 0.906 \\
R$_{\text{own}}$=7 & P$_{\text{dinam}}$ & 18 & 0.609 & \textbf{0.421} & 2 & 0.286 & 0.491 & 0.528 & 158 & 0.929 \\
R$_{\text{ref}}$=170 & P$_{\text{clr}}$ & 53 & 0.603 & 0.471 & 2 & 0.286 & 0.445 & 0.541 & 147 & 0.865 \\
 & P$_{\text{com}}$ & 267 & 0.632 & 0.492 & 2 & 0.286 & 0.460 & 0.521 & 128 & 0.753 \\
 & P$_{\text{per-clr}}$ & 186 & \textbf{0.649} & 0.447 & \textbf{0} & \textbf{0.000} & 0.450 & \textbf{0.507} & 138 & 0.812 \\
 & P$_{\text{per-com}}$ & 232 & 0.636 & 0.466 & \textbf{0} & \textbf{0.000} & 0.464 & 0.510 & \textbf{127} & \textbf{0.747} \\
\addlinespace
CV$\rightarrow$EU & P$_{\text{SVO}}$ & 51 & 0.511 & 0.520 & 9 & 0.643 & 0.491 & 0.491 & 148 & 0.871 \\
R$_{\text{own}}$=14 & P$_{\text{dinam}}$ & 25 & 0.566 & \textbf{0.445} & 5 & 0.357 & \textbf{0.591} & 0.458 & 150 & 0.882 \\
R$_{\text{ref}}$=170 & P$_{\text{clr}}$ & 75 & 0.573 & 0.483 & 5 & 0.357 & 0.527 & 0.462 & 134 & 0.788 \\
 & P$_{\text{com}}$ & 829 & 0.591 & 0.520 & 2 & 0.143 & 0.542 & 0.489 & 101 & 0.594 \\
 & P$_{\text{per-clr}}$ & 173 & 0.581 & 0.485 & 3 & 0.214 & 0.511 & \textbf{0.451} & 114 & 0.671 \\
 & P$_{\text{per-com}}$ & 614 & \textbf{0.601} & 0.491 & \textbf{1} & \textbf{0.071} & 0.553 & 0.456 & \textbf{67} & \textbf{0.394} \\
\addlinespace
HP$\rightarrow$EU & P$_{\text{SVO}}$ & 65 & 0.548 & 0.475 & 3 & 0.273 & 0.516 & 0.472 & 139 & 0.818 \\
R$_{\text{own}}$=11 & P$_{\text{dinam}}$ & 72 & 0.615 & \textbf{0.436} & 1 & 0.091 & \textbf{0.566} & \textbf{0.435} & 129 & 0.759 \\
R$_{\text{ref}}$=170 & P$_{\text{clr}}$ & 220 & 0.600 & 0.488 & 2 & 0.182 & 0.517 & 0.462 & 117 & 0.688 \\
 & P$_{\text{com}}$ & 1046 & 0.610 & 0.513 & \textbf{0} & \textbf{0.000} & 0.562 & 0.487 & \textbf{75} & \textbf{0.441} \\
 & P$_{\text{per-clr}}$ & 428 & \textbf{0.624} & 0.475 & \textbf{0} & \textbf{0.000} & 0.556 & 0.449 & 98 & 0.576 \\
 & P$_{\text{per-com}}$ & 506 & 0.604 & 0.482 & \textbf{0} & \textbf{0.000} & 0.551 & 0.459 & 94 & 0.553 \\
\addlinespace
NM$\rightarrow$EU & P$_{\text{SVO}}$ & 18 & 0.509 & 0.462 & 13 & 0.765 & \textbf{0.521} & 0.515 & 155 & 0.912 \\
R$_{\text{own}}$=17 & P$_{\text{dinam}}$ & 2 & 0.529 & 0.491 & 16 & 0.941 & 0.503 & 0.572 & 168 & 0.988 \\
R$_{\text{ref}}$=170 & P$_{\text{clr}}$ & 6 & 0.507 & 0.465 & 15 & 0.882 & 0.484 & 0.568 & 167 & 0.982 \\
 & P$_{\text{com}}$ & 173 & 0.571 & 0.489 & 4 & 0.235 & 0.443 & 0.511 & 129 & 0.759 \\
 & P$_{\text{per-clr}}$ & 13 & 0.513 & \textbf{0.451} & 14 & 0.824 & 0.510 & 0.525 & 159 & 0.935 \\
 & P$_{\text{per-com}}$ & 160 & \textbf{0.602} & 0.453 & \textbf{3} & \textbf{0.176} & 0.478 & \textbf{0.485} & \textbf{113} & \textbf{0.665} \\
\addlinespace
PN$\rightarrow$EU & P$_{\text{SVO}}$ & 182 & 0.553 & 0.464 & 45 & 0.608 & 0.495 & 0.481 & 123 & 0.724 \\
R$_{\text{own}}$=74 & P$_{\text{dinam}}$ & 47 & 0.565 & \textbf{0.426} & 53 & 0.716 & \textbf{0.572} & \textbf{0.457} & 147 & 0.865 \\
R$_{\text{ref}}$=170 & P$_{\text{clr}}$ & 141 & 0.516 & 0.463 & 45 & 0.608 & 0.487 & 0.476 & 130 & 0.765 \\
 & P$_{\text{com}}$ & 824 & 0.576 & 0.495 & 22 & 0.297 & 0.536 & 0.496 & \textbf{90} & \textbf{0.529} \\
 & P$_{\text{per-clr}}$ & 231 & 0.546 & 0.449 & 40 & 0.541 & 0.504 & 0.464 & 109 & 0.641 \\
 & P$_{\text{per-com}}$ & 392 & \textbf{0.577} & 0.462 & \textbf{19} & \textbf{0.257} & 0.535 & 0.466 & 97 & 0.571 \\
\addlinespace
UK$\rightarrow$EU & P$_{\text{SVO}}$ & 32 & 0.478 & 0.493 & 24 & 0.750 & 0.495 & 0.498 & 154 & 0.906 \\
R$_{\text{own}}$=32 & P$_{\text{dinam}}$ & 18 & 0.533 & \textbf{0.446} & 24 & 0.750 & \textbf{0.552} & 0.478 & 155 & 0.912 \\
R$_{\text{ref}}$=170 & P$_{\text{clr}}$ & 56 & 0.532 & 0.467 & 17 & 0.531 & 0.521 & 0.479 & 141 & 0.829 \\
 & P$_{\text{com}}$ & 333 & 0.578 & 0.483 & 9 & 0.281 & 0.516 & 0.483 & 109 & 0.641 \\
 & P$_{\text{per-clr}}$ & 136 & 0.562 & 0.461 & 8 & 0.250 & 0.507 & 0.470 & 120 & 0.706 \\
 & P$_{\text{per-com}}$ & 342 & \textbf{0.590} & 0.459 & \textbf{3} & \textbf{0.094} & 0.530 & \textbf{0.461} & \textbf{91} & \textbf{0.535} \\
\bottomrule
\end{tabular}
\caption{Full evaluation table across all datasets and single-runs per pipeline. P$_{\text{dinam}}$ is our re-implementation of the original DiNaM pipeline, included for comparison (App.~\ref{sec:harmonize}). Bold marks the best value per dataset and column.}
\label{tab:all_pipelines_collapse}
\end{table*}

\begin{table*}[h]
\centering
\small
\begin{tabular}{ll r rrrr rrrr}
\toprule
 & & \# Narrative & \multicolumn{4}{c}{Recovery} & \multicolumn{4}{c}{Mining} \\
\cmidrule(lr){4-7} \cmidrule(lr){8-11}
DS & Method & Labels & Hung.$\uparrow$ & WCD$\downarrow$ & Coll.$\downarrow$ & C/R$\downarrow$ & Hung.$\uparrow$ & WCD$\downarrow$ & Coll.$\downarrow$ & C/R$\downarrow$ \\
\midrule
EU & P$_{\text{SVO}}$ & 1341 & 0.571 & 0.495 & 87 & 0.512 & -- & -- & -- & -- \\
R$_{\text{own}}$=170 & P$_{\text{dinam}}$ & 354 & \textbf{0.593} & \textbf{0.416} & 83 & 0.488 & -- & -- & -- & -- \\
 & P$_{\text{clr}}$ & 1058 & 0.566 & 0.485 & 82 & 0.482 & -- & -- & -- & -- \\
 & P$_{\text{com}}$ & 745 & 0.524 & 0.505 & 109 & 0.641 & -- & -- & -- & -- \\
 & P$_{\text{per-clr}}$ & 1791 & 0.590 & 0.473 & \textbf{66} & \textbf{0.388} & -- & -- & -- & -- \\
 & P$_{\text{per-com}}$ & 1266 & 0.566 & 0.481 & 73 & 0.429 & -- & -- & -- & -- \\
\addlinespace
CO$\rightarrow$CARDS2 & P$_{\text{SVO}}$ & 27 & 0.586 & 0.468 & 2 & 0.286 & 0.462 & 0.497 & 29 & 0.906 \\
R$_{\text{own}}$=7 & P$_{\text{dinam}}$ & 18 & 0.609 & \textbf{0.421} & 2 & 0.286 & 0.501 & 0.483 & 23 & 0.719 \\
R$_{\text{ref}}$=32 & P$_{\text{clr}}$ & 53 & 0.603 & 0.471 & 2 & 0.286 & 0.460 & 0.517 & \textbf{18} & \textbf{0.562} \\
 & P$_{\text{com}}$ & 267 & 0.632 & 0.492 & 2 & 0.286 & 0.515 & 0.512 & 21 & 0.656 \\
 & P$_{\text{per-clr}}$ & 186 & \textbf{0.649} & 0.447 & \textbf{0} & \textbf{0.000} & 0.517 & \textbf{0.477} & 23 & 0.719 \\
 & P$_{\text{per-com}}$ & 232 & 0.636 & 0.466 & \textbf{0} & \textbf{0.000} & \textbf{0.534} & 0.480 & 27 & 0.844 \\
\addlinespace
DiNaM$\rightarrow$DGT & P$_{\text{SVO}}$ & 1341 & -- & -- & -- & -- & 0.570 & 0.496 & 82 & 0.494 \\
R$_{\text{ref}}$=166 & P$_{\text{dinam}}$ & 354 & -- & -- & -- & -- & \textbf{0.599} & \textbf{0.413} & 73 & 0.440 \\
 & P$_{\text{clr}}$ & 1058 & -- & -- & -- & -- & 0.568 & 0.484 & 76 & 0.458 \\
 & P$_{\text{com}}$ & 745 & -- & -- & -- & -- & 0.526 & 0.503 & 110 & 0.663 \\
 & P$_{\text{per-clr}}$ & 1791 & -- & -- & -- & -- & 0.592 & 0.472 & \textbf{62} & \textbf{0.373} \\
 & P$_{\text{per-com}}$ & 1266 & -- & -- & -- & -- & 0.569 & 0.479 & 74 & 0.446 \\
\bottomrule
\end{tabular}
\caption{Additional evaluation table across specific settings. DiNaM$\rightarrow$DGT evaluates the DiNaM dataset (scraped based on instructions in \url{https://github.com/wsosnowski/DiNaM}) against the ground-truth labels used in its original implementation. Bold marks the best value per dataset/column.}
\label{tab:all_pipelines_collapse_selected}
\end{table*}

\clearpage
\section{Metric Correlations}
\label{sec:metrics-discussion}

\begin{table}[h]
\centering
\small
\begin{tabular}{lrrrrr}
\toprule
\multicolumn{6}{l}{\textit{Recovery, $n{=}48$}} \\
\addlinespace[2pt]
       & $|L|$ & Hung  & WCD   & Collapse & C/R  \\
$|L|$    & ---   & 0.52  & 0.46  & -0.08    & -0.53 \\
Hung     &       & ---   & 0.02  & -0.72    & -0.85 \\
WCD      &       &       & ---   & -0.10    & -0.15 \\
Collapse &       &       &       & ---      & 0.82 \\
\midrule
\multicolumn{6}{l}{\textit{Mining, $n{=}54$}} \\
\addlinespace[2pt]
       & $|L|$ & Hung  & WCD   & Collapse & C/R  \\
$|L|$    & ---   & 0.52  & -0.25 & -0.67    & -0.90 \\
Hung     &       & ---   & -0.60 & -0.34    & -0.52 \\
WCD      &       &       & ---   & 0.25     & 0.36 \\
Collapse &       &       &       & ---      & 0.78 \\
\bottomrule
\end{tabular}
\caption{Spearman $\rho$ between metrics, computed separately on the recovery and mining sides of both evaluation tables (Tables ~\ref{tab:all_pipelines_collapse} and ~\ref{tab:all_pipelines_collapse_selected}). \emph{Recovery} excludes the CO$\rightarrow$CARDS2 row, whose recovery cells duplicate CO ($n{=}48$). \emph{Mining} corresponds to the \emph{Mining} columns of Tab.~\ref{tab:all_pipelines_collapse} ($n{=}54$). $|L|$ is the number of NLs produced by a pipeline.}
\label{tab:metrics_stat_a}
\end{table}

We report Spearman correlations on the recovery and mining sides of Tables~\ref{tab:all_pipelines_collapse} and~\ref{tab:all_pipelines_collapse_selected} separately ($n{=}48$ recovery, $n{=}54$ mining. $|L|$ denotes the number of NLs produced by a pipeline). Cells are not independent (multiple pipelines per dataset and vice versa). We therefore read correlations as directional summaries and not as hypothesis tests. We interpret sign and relative magnitude, not precise values, and use the bootstrap CIs in App.~\ref{app:bootstrap-cis} to identify pairs whose direction is unstable.

\paragraph{Among metrics.}
\begin{itemize}[nosep]
    \item \textbf{Collapse and C/R} are tightly coupled ($+0.82$ rec / $+0.78$ min): C/R $= \mathrm{Coll}/m$, so the two carry largely overlapping information.
    \item \textbf{Hungarian and Collapse/C/R} are negatively correlated on both sides ($-0.72, -0.85$ rec; $-0.34, -0.52$ min). Hungarian agrees directionally with both Collapse and C/R, but more strongly on recovery than on mining (where the mining CIs come closer to zero).
    \item \textbf{Hungarian and WCD} are uncorrelated on recovery ($+0.02$, CI $[-0.26, +0.34]$) but agree on mining ($-0.60$, CI $[-0.73, -0.40]$;  Hungarian $\uparrow$ and WCD $\downarrow$, so a negative correlation means agreement). The reason is the number of generated labels relative to the number of references, $|L|/m$. WCD averages, over all $|L|$ NLs, the distance to the nearest reference, so every extra label that matches no reference makes WCD worse. Hungarian scores only the best $\min(m,n)$ one-to-one pairs, so once $|L| \ge m$, extra labels can only improve it. On recovery, $|L|/m$ is large (median $7.5$, range $0.1$--$129$) and the two metrics respond to label count in opposite directions: P$_{\text{dinam}}$, which generates the fewest labels on six of the seven datasets, has the best WCD on six of them, while the best Hungarian on every dataset goes to one of the two Persona pipelines, which generate many times more labels. On mining, $|L|/m$ is close to 1 (median $1.4$), the size effect disappears, and the two metrics agree.
\end{itemize}
The remaining pairs (WCD $\times$ Collapse, WCD $\times$ C/R) fall within $\pm 0.36$ on both sides with bootstrap CIs spanning zero on at least one side. Reported in Tab.~\ref{tab:metrics_stat_a} for completeness.

\paragraph{Versus NL count.}
$|L|$ correlates with Hungarian, WCD and C/R, asymmetrically across sides, and with Collapse on mining only (full numbers in Tab.~\ref{tab:metrics_stat_a}). The strong $|L| \times \mathrm{C/R}$ correlation on mining ($-0.90$) is partly mechanical: when $|L| < m$, Collapse has a floor of $m - |L|$, so additional NLs strictly reduce the C/R floor. The same effect is muted on recovery, where $m \le 32$ for most datasets and $|L|$ typically exceeds $m$.

\paragraph{Practical implication.}
Hungarian is the most $|L|$-robust of the four and is what we rely on for cross-dataset pipeline comparisons whose $|L|$ differ substantially. WCD is retained for comparability with prior work \citep{2025dinam} but is $|L|$-sensitive on recovery. Collapse explicitly counts mode collisions that Hungarian and WCD smooth over. C/R normalizes Collapse for cross-taxonomy comparison. Within-dataset pipeline comparisons (fixed $m$) are robust under all four metrics.

\section{Automated Metrics Correlations: Bootstrap Confidence Intervals}
\label{app:bootstrap-cis}

Tab.~\ref{tab:bootstrap-cis} reports $95\%$ bootstrap confidence intervals (percentile method, $1000$ resamples with replacement over the
cells of each side) for the Spearman correlations summarized in Tab.~\ref{tab:metrics_stat_a}.

\begin{table}[H]
\centering\small
\setlength{\tabcolsep}{4pt}
\caption{Bootstrap $95\%$ CIs for the Spearman correlations reported in Tab.~\ref{tab:metrics_stat_a} (full evaluation table: Tab.~\ref{tab:all_pipelines_collapse}); $1000$ resamples, percentile method.}
\label{tab:bootstrap-cis}
\begin{tabular}{lcc}
\toprule
Pair & Recovery ($n{=}48$) & Mining ($n{=}54$) \\
\midrule
$|L|\times$ Hungarian       & $[+0.26,+0.72]$ & $[+0.26,+0.72]$ \\
$|L|\times$ WCD             & $[+0.19,+0.67]$ & $[-0.49,+0.05]$ \\
$|L|\times$ Collapse        & $[-0.40,+0.23]$ & $[-0.87,-0.40]$ \\
$|L|\times$ C/R             & $[-0.74,-0.24]$ & $[-0.95,-0.80]$ \\
\addlinespace
Hungarian $\times$ WCD      & $[-0.26,+0.34]$ & $[-0.73,-0.40]$ \\
Hungarian $\times$ Collapse & $[-0.82,-0.56]$ & $[-0.61,-0.06]$ \\
Hungarian $\times$ C/R      & $[-0.92,-0.72]$ & $[-0.73,-0.26]$ \\
\addlinespace
WCD $\times$ Collapse       & $[-0.40,+0.19]$ & $[-0.07,+0.51]$ \\
WCD $\times$ C/R            & $[-0.44,+0.14]$ & $[+0.09,+0.59]$ \\
\addlinespace
Collapse $\times$ C/R       & $[+0.69,+0.89]$ & $[+0.57,+0.95]$ \\
\bottomrule
\end{tabular}
\end{table}

\section{Robustness of Singleton Findings}
\label{sec:robustness-singletons}
To rule out alternative explanations of the singleton finding, we verify it along four axes.

\emph{Seed stability (Leiden).} We re-run Leiden community detection under 10 independent random seeds with all LLM outputs fixed. Singleton counts are nearly deterministic: $P_{\text{per-com}}$ produces $70.1 \pm 0.3$ singletons on CO and $148.6 \pm 4.2$ on PN. $P_{\text{com}}$ produces $476.2 \pm 15.3$ on PN. The standard deviation is at most $3.2\%$ of the mean in every case.
\emph{Pipeline transferability.} The singleton phenomenon is not specific to the persona pipeline. On PN, $P_{\text{com}}$ (where no personas are involved) yields 507/824 singleton narrative labels (61.5\%), exceeding $P_{\text{per-com}}$'s 149/392 (38.0\%). 

\emph{Persona ablation.} Across 15 persona-subset configurations (1--4 personas) $\times$ 10 Leiden seeds on PN, mean singleton counts remain in 150--154 and the singleton fraction of communities in 58--71\% regardless of subset (§\ref{sec:singleton} reports shares of NLs, which are lower because a community can yield several NLs). Per-subset seed variance is 1--4 singletons. Cross-subset variance is 10--75. 

\emph{Upstream LLM variance.} We re-ran the full pipeline three times on PN (three end-to-end replicate runs, $n{=}3$) with the same Gemma-4-31B-it \cite{gemma_4_31b} model at $\text{temp}{=}0$ and the same Leiden seed, varying only the input-document order so that vLLM's continuous batching produces a different batch composition each replicate. With \texttt{concurrency=100} and the fused-attention kernels used at $\text{temp}{=}0$, vLLM is not bit-deterministic across runs. Stage-1 (claim extraction) is reproducible to $0.2\%$ across replicates (19{,}349 / 19{,}385 / 19{,}390 claims), and the post-extraction graphs match within $1\%$ in both node and edge counts.
Singleton counts at the production Leiden seed are 140 / 150 / 159 (mean 149.7, stdev 9.5 over 3 replicates). Within each replicate, additionally varying the Leiden seed over 10 values yields 133--174 singletons (3 replicates $\times$ 10 Leiden seeds = 30 measurements; mean 154.4, stdev 11.7). The Leiden contribution to total variance (within-replicate stdev $\approx 6$, well-estimated from 30 samples) is smaller than the apparent LLM contribution (between-replicate stdev $\approx 11$, estimated from only 3 replicates). The singleton fraction of communities, however, is stable at 60.6 / 59.8 / 62.8\% across replicates.

\section{Parameter Sweep Analysis and Topical Coverage}
\label{app:topic}

To characterize the topical distribution of narrative labels (NL) across the five pipelines on PN, we apply zero-shot LLM annotation with an explicit four-way label set: \emph{War in Ukraine}, \emph{Climate}, \emph{Other}, \emph{Noise}. The annotator model is Gemma-4-31B-it.

We use a zero-shot prompt with explicit label definitions and a borderline-case decision rule, in three variants that differ in label order and answer format, and classify each NL five times per variant at sampling temperature $t{=}0.5$ (self-consistency, $k{=}5$). The majority label within each variant is kept, and the final label is the majority across the three variants. The results are documented in Table~\ref{tab:topical_coverage}.
The default-parameter rows of Table~\ref{tab:sweep-pn} are the runs the annotators of §\ref{sec:openworld} saw; their observation that the P$_{\text{per-clr}}$ output was almost entirely about the war in Ukraine (§\ref{sec:param-sweep}) agrees with the LLM annotation.

\begin{table*}[h]
\centering
\small
\begin{tabular}{lrrrrrrrrrrrrr}
\toprule
\multicolumn{1}{c}{} & \multicolumn{1}{c}{} & \multicolumn{4}{c}{Topic \%} & \multicolumn{4}{c}{Recovery (own ref)} &
\multicolumn{4}{c}{Mining (EU)} \\
\cmidrule(lr){3-6}\cmidrule(lr){7-10}\cmidrule(lr){11-14}
Param & $N$ & ukr & cli & oth & noi & H$\uparrow$ & WCD$\downarrow$ & C$\downarrow$ & C/R$\downarrow$ & H$\uparrow$ & WCD$\downarrow$ & C$\downarrow$ & C/R$\downarrow$ \\
\midrule
\multicolumn{14}{l}{\textit{HDBSCAN sweep on P$_{\text{clr}}$, (mcs, ms) parameter pair, ascending mcs}} \\
\addlinespace[2pt]
(2,1) & 4069 & 55.8 & 27.3 & 16.9 & 0.1 & 0.690 & 0.394 & 13 & 0.176 & 0.664 & 0.370 & 78 & 0.459 \\
(5,2) & 1300 & 52.1 & 29.5 & 18.3 & 0.1 & 0.673 & 0.394 & 25 & 0.338 & 0.640 & 0.371 & 88 & 0.518 \\
(10,5) & 566 & 53.4 & 30.4 & 16.2 & 0.0 & 0.655 & 0.390 & 25 & 0.338 & 0.621 & 0.369 & 105 & 0.618 \\
(15,10) & 332 & 52.1 & 33.7 & 14.2 & 0.0 & 0.645 & 0.390 & 34 & 0.459 & 0.599 & 0.375 & 114 & 0.671 \\
(25*,20) & 141 & 83.0 & 2.1 & 14.9 & 0.0 & 0.582 & 0.384 & 51 & 0.689 & 0.576 & 0.381 & 128 & 0.753 \\
\midrule
\multicolumn{14}{l}{\textit{HDBSCAN sweep on P$_{\text{per-clr}}$, (mcs, ms) parameter pair, ascending mcs}} \\
\addlinespace[2pt]
(2,1) & 4439 & 58.9 & 25.9 & 15.2 & 0.0 & 0.698 & 0.380 & 14 & 0.189 & 0.674 & 0.358 & 74 & 0.435 \\
(5,2) & 1982 & 57.3 & 26.7 & 16.0 & 0.0 & 0.683 & 0.382 & 21 & 0.284 & 0.660 & 0.359 & 79 & 0.465 \\
(10,5) & 938 & 60.5 & 24.7 & 14.7 & 0.0 & 0.668 & 0.381 & 31 & 0.419 & 0.639 & 0.359 & 92 & 0.541 \\
(15,10) & 508 & 61.2 & 25.0 & 13.8 & 0.0 & 0.657 & 0.379 & 29 & 0.392 & 0.620 & 0.365 & 93 & 0.547 \\
(25*,20) & 231 & 85.7 & 1.7 & 12.6 & 0.0 & 0.608 & 0.374 & 41 & 0.554 & 0.582 & 0.373 & 114 & 0.671 \\
\midrule
\multicolumn{14}{l}{\textit{Leiden modularity-$\gamma$ sweep on P$_{\text{com}}$ (RBConfiguration), $\gamma$ ascending}} \\
\addlinespace[2pt]
0.1 & 558 & 19.7 & 36.0 & 40.5 & 3.8 & 0.631 & 0.444 & 44 & 0.595 & 0.596 & 0.414 & 123 & 0.724 \\
0.25 & 802 & 27.8 & 33.8 & 35.5 & 2.9 & 0.646 & 0.438 & 41 & 0.554 & 0.613 & 0.409 & 106 & 0.624 \\
0.5 & 809 & 30.8 & 31.5 & 34.6 & 3.1 & 0.652 & 0.434 & 36 & 0.486 & 0.614 & 0.407 & 106 & 0.624 \\
0.75 & 930 & 39.8 & 27.1 & 30.6 & 2.5 & 0.652 & 0.430 & 33 & 0.446 & 0.618 & 0.404 & 95 & 0.559 \\
1$^{*}$ & 824 & 38.1 & 28.8 & 30.3 & 2.8 & 0.651 & 0.429 & 30 & 0.405 & 0.618 & 0.400 & 104 & 0.612 \\
1.5 & 896 & 42.3 & 27.9 & 27.5 & 2.3 & 0.655 & 0.424 & 34 & 0.459 & 0.620 & 0.398 & 89 & 0.524 \\
2 & 998 & 43.0 & 29.1 & 25.9 & 2.1 & 0.657 & 0.421 & 33 & 0.446 & 0.623 & 0.394 & 93 & 0.547 \\
5 & 1362 & 45.7 & 31.3 & 21.5 & 1.5 & 0.668 & 0.416 & 24 & 0.324 & 0.628 & 0.388 & 91 & 0.535 \\
10 & 1938 & 49.9 & 29.8 & 18.9 & 1.3 & 0.671 & 0.411 & 26 & 0.351 & 0.635 & 0.384 & 85 & 0.500 \\
\midrule
\multicolumn{14}{l}{\textit{Leiden modularity-$\gamma$ sweep on P$_{\text{per-com}}$ (RBConfiguration), $\gamma$ ascending}} \\
\addlinespace[2pt]
0.1 & 224 & 22.8 & 33.5 & 43.3 & 0.4 & 0.628 & 0.402 & 46 & 0.622 & 0.583 & 0.389 & 113 & 0.665 \\
0.25 & 522 & 32.0 & 29.3 & 38.5 & 0.2 & 0.651 & 0.404 & 35 & 0.473 & 0.623 & 0.378 & 110 & 0.647 \\
0.5 & 456 & 38.4 & 29.6 & 32.0 & 0.0 & 0.656 & 0.393 & 35 & 0.473 & 0.623 & 0.373 & 102 & 0.600 \\
0.75 & 381 & 38.3 & 35.2 & 26.5 & 0.0 & 0.655 & 0.392 & 39 & 0.527 & 0.612 & 0.376 & 112 & 0.659 \\
1$^{*}$ & 392 & 37.0 & 37.5 & 25.3 & 0.3 & 0.655 & 0.390 & 34 & 0.459 & 0.617 & 0.370 & 113 & 0.665 \\
1.5 & 522 & 43.9 & 33.1 & 22.8 & 0.2 & 0.665 & 0.385 & 27 & 0.365 & 0.629 & 0.362 & 107 & 0.629 \\
2 & 647 & 45.4 & 34.0 & 20.4 & 0.2 & 0.667 & 0.385 & 31 & 0.419 & 0.630 & 0.364 & 110 & 0.647 \\
5 & 1121 & 50.7 & 31.6 & 17.8 & 0.0 & 0.680 & 0.379 & 24 & 0.324 & 0.641 & 0.358 & 99 & 0.582 \\
10 & 1673 & 54.5 & 29.6 & 15.9 & 0.1 & 0.685 & 0.379 & 21 & 0.284 & 0.656 & 0.356 & 90 & 0.529 \\
\bottomrule
\end{tabular}
\caption{PN clustering parameter sweep. Topic \%: share of NLs about the war in Ukraine (ukr), climate (cli), other topics (oth), or noise (noi), per App.~\ref{app:topic}. Recovery: NLs vs.\ PN reference taxonomy. Mining: NLs vs.\ EU DisinfoTest taxonomy. H = Hungarian, C = Collapse; WCD, C, and C/R lower=better, H higher=better. $^{*}$ marks the default parameter value of the underlying algorithm (HDBSCAN: $\text{mcs}{=}25$, $\text{ms}{=}20$; Leiden modularity-$\gamma$ with the RBConfiguration quality function of \texttt{leidenalg}: $\gamma{=}1.0$).
}
\label{tab:sweep-pn}
\end{table*}

\begin{table*}[h]
\centering
\small
\setlength{\tabcolsep}{6pt}
\begin{tabular}{lr rrrr rr}
\toprule
& & \multicolumn{4}{c}{Topic share (\%)} & \multicolumn{2}{c}{Robustness} \\
\cmidrule(lr){3-6}\cmidrule(lr){7-8}
Pipeline & $L$ & Ukraine & Climate & Other & Noise & 3-var.\ agree & mean $k{=}5$ agree \\
\midrule
P$_{\text{SVO}}$        & 182 &  5.5 &  5.0 &  2.8 & \textbf{86.8} & 98.3 & 1.000 \\
P$_{\text{clr}}$      & 141 & \textbf{83.0} &  2.1 & 14.9 &  0.0 & 95.7 & 1.000 \\
P$_{\text{com}}$            & 824 & 38.1 & 28.8 & 30.3 &  2.8          & 95.3 & 0.997 \\
P$_{\text{per-clr}}$     & 231 & \textbf{85.7} &  1.7 & 12.6 &  0.0 & 94.8 & 0.998 \\
P$_{\text{per-com}}$  & 392 & 37.0 & 37.5 & 25.3 &  0.3          & 95.9 & 0.998 \\
\bottomrule
\end{tabular}
\caption{Per-pipeline topical distribution of narrative labels on PN.}
\label{tab:topical_coverage}
\end{table*}

\clearpage
\section{Superclaim-Oriented Prompt}
\label{sec:prompt}
\noindent ``Superclaim'' is a formulation from \citet{Coan2021}, which addresses the idea of grouping together disinformation claims. Subsequent datasets like PolyNarrative use narrative-oriented terms for this purpose.

\begin{tcolorbox}[colback=gray!5, colframe=gray!50, boxrule=0.5pt]

Below is a set of claims from the same thematic cluster. Your task is to \
synthesize them into 1-3 super-claims that capture the cluster's core message.

A super-claim is a higher-level claim that groups together multiple specific 
claims. It is a normalized, declarative assertion — a statement asserting that 
something is the case, which one could agree or disagree with.

Important: a super-claim is NOT a newspaper headline, NOT a neutral topic 
label, and NOT a verbose summary. It is a concise, argumentative statement 
that captures what is being argued.

For each super-claim, provide:\\
- "title": The super-claim itself, formulated as a single normalized 
declarative assertion (one sentence, 10-20 words). It must read as a claim, 
not as a headline or topic.\\
- "description": A slightly fuller restatement of the same claim with 
additional context (1-2 sentences). Also formulated as an assertion, not as 
a summary of what texts argue.

Do not simply list all claims. Synthesize them into higher-level assertions. 
If multiple claims argue the same point from different angles, combine them 
into one super-claim.

Claims in this cluster:
{content}

\end{tcolorbox}

\clearpage

\section{Computational Setup and Budget}
\label{app:compute}

All experiments run on a single NVIDIA H100 80GB GPU. Gemma-4-31B-it is served via vLLM. The topical-coverage LLM annotation (App.~\ref{app:topic}) uses $t{=}0.5$ with $k{=}5$ self-consistency. Qwen3-Embedding-4B (pipelines) and harrier-oss-v1-0.6b (closed-world metrics, §\ref{sec:metrics}) run on the same machine. The LLM-judge runs (App.~\ref{sec:llm-judge}) reuse this setup with Qwen3.5-27B \citep{qwen2026qwen35} as the second judge ensemble.

On the PolyNarrative dataset, for example, mean wall-clock per run is 33 min for $P_{\text{per-com}}$ and 16 min for $P_{\text{com}}$. Clustering-only pipelines complete in under 5 min.
All pipeline experiments (App.~Tables~\ref{tab:all_pipelines_collapse} and~\ref{tab:all_pipelines_collapse_selected}) total at about 100 H100-hours.
Parameter-sweep and singleton-robustness experiments (App.~\ref{app:topic}, App.~\ref{sec:robustness-singletons}. 174 sweep runs plus 3 PN replicates) add about 30 H100-hours. Combined compute for experiments reported in this paper is roughly 130 H100-hours on a single GPU.

\section{Harmonized Pipeline Configuration}
\label{sec:harmonize}
All pipelines run on Gemma-4-31B-it (open-weight; replacing the proprietary models used by \citet{2025dinam} and \citet{xu2025noderagstructuringgraphbasedrag} for reproducibility) and Qwen3-Embedding-4B (replacing DiNaM's monolingual SFR). LLM-based pipelines share a single superclaim NL prompt (documented in App.~\ref{sec:prompt}), which prompts for a declarative assertion-style narrative labels and thus produces labels closer to the formulations in the taxonomies. While DiNaM's summary prompt also produces good results and in fact outperforms its harmonized counterpart P$_{\text{clr}}$ on WCD and Hungarian (6/7 datasets each), the +SC variants yield better results on Collapse (see Table~\ref{tab:all_pipelines_collapse}). We speculate that this has to do with specific failure modes of the DiNaM prompt, in some cases producing labels like ``Use specific supplements, detoxes, and alternative medicines to cure diseases and treat COVID-19'' instead of declarative assertions like ``COVID-19 and other severe illnesses can be effectively treated with alternative medications and specific nutrient protocols'', produced with our +SC variant. Since our goal is to enable human validation of these outputs, we opt for our prompt due to its performance on the Collapse metrics and the generated formulation style being more aligned with existing taxonomies.

\clearpage
\section{Human Validation}

\subsection{Inter-annotator Agreement Evaluation}
\label{app:iaa}
\begin{strip}
  \centering\small
  \setlength{\tabcolsep}{3pt}
    \captionof{table}{Inter-annotator agreement across the four runs (annotators $a$, $b$). \textbf{All}: every row. \textbf{Lo}/\textbf{Hi}: rows where \texttt{confidence}~$\in\{1,2,3\}$ / $\{4,5\}$. Pairwise $\kappa$ and agreement require both annotators in band on the same row. Dashes and degenerate values (exact $0$ or $1$) arise because annotator $a$ filled \texttt{confidence} only on \emph{yes} rows. \texttt{is\_narr}: \texttt{is\_disinfo\_narrative} (yes / no / unclear). \texttt{match\_coll}: \texttt{label\_match} with the four retrieved labels collapsed into one  class. A \texttt{label\_match} answer counts only on \emph{yes} rows. Per-annotator rates share denominator $N$. Set-overlap rows give $|a\cap b|/|a\cup b|$ as NL counts.}
  \label{tab:iaa-all}
  \begin{tabular}{l l ccc ccc ccc ccc}
  \toprule
  & & \multicolumn{6}{c}{CO} & \multicolumn{6}{c}{PN} \\
  \cmidrule(lr){3-8}\cmidrule(lr){9-14}
  & & \multicolumn{3}{c}{P$_{\text{per-clr}}$} & \multicolumn{3}{c}{P$_{\text{per-com}}$}
    & \multicolumn{3}{c}{P$_{\text{per-clr}}$} & \multicolumn{3}{c}{P$_{\text{per-com}}$} \\
  \cmidrule(lr){3-5}\cmidrule(lr){6-8}\cmidrule(lr){9-11}\cmidrule(lr){12-14}
  & & All & Lo & Hi & All & Lo & Hi & All & Lo & Hi & All & Lo & Hi \\
  \midrule
  \multicolumn{14}{l}{\textit{Per-annotator label rate \% (denom $=N$ rows)}} \\
  \addlinespace[2pt]
  $a$ & yes\%   & 66.1 & 30.1 & 36.0 & 58.2 & 28.9 & 29.3 & 66.2 & 26.8 & 39.0 & 54.3 & 13.8 & 39.0 \\
  $b$ & yes\%   & 58.1 &  8.6 & 48.9 & 51.7 &  6.0 & 44.8 & 52.8 &  6.5 & 46.3 & 40.3 &  4.8 & 35.5 \\
  \addlinespace[2pt]
  $a$ & other\% & 14.5 &  9.7 &  4.8 & 17.2 &  9.9 &  7.3 & 20.3 & 10.8 &  9.5 & 12.8 &  4.8 &  7.9 \\
  $b$ & other\% &  6.5 &  2.7 &  3.8 & 15.5 &  2.2 & 13.4 & 24.7 &  4.8 & 19.9 & 12.2 &  1.5 & 10.7 \\
  \addlinespace[2pt]
  $a$ & none\%  &  0.0 &  0.0 &  0.0 &  1.3 &  0.9 &  0.4 & 12.1 &  4.3 &  7.8 & 10.5 &  1.3 &  7.7 \\
  $b$ & none\%  &  0.0 &  0.0 &  0.0 &  2.6 &  0.4 &  2.2 &  5.2 &  0.0 &  5.2 &  6.6 &  1.3 &  5.4 \\
  \midrule
  \multicolumn{14}{l}{\textit{Cohen $\kappa$ ($a$ vs $b$, paired rows in band)}} \\
  \addlinespace[2pt]
  \multicolumn{2}{l}{$\kappa$ \texttt{is\_narr}}    & 0.497 & --        & --    & 0.564 & --        & --    & 0.554 & --    & --    & 0.578 &
  --    & --    \\
  \multicolumn{2}{l}{$\kappa$ \texttt{match\_coll}} & 0.437 & $-$0.154  & 0.143 & 0.507 & $-$0.143  & 0.218 & 0.528 & 0.074 & 0.494 & 0.540 &
  0.237 & 0.344 \\
  \multicolumn{2}{l}{$\kappa$ \texttt{merged}}      & 0.558 & 1.000     & 1.000 & 0.651 & 0.000     & 1.000 & 0.610 & 0.138 & 0.598 & 0.621 &
  0.294 & 0.330 \\
  \multicolumn{2}{l}{$\kappa$ \texttt{label\_match}}& 0.391 & 0.167     & 0.465 & 0.446 & 0.077     & 0.326 & 0.517 & 0.059 & 0.549 & 0.510 &
  0.250 & 0.426 \\
  \midrule
  \multicolumn{14}{l}{\textit{Raw agreement \% ($a$ vs $b$)}} \\
  \addlinespace[2pt]
  \multicolumn{2}{l}{agree \texttt{is\_narr}}    & 74.2 & --    & --    & 76.3 & --    & --    & 76.6 & --    & --    & 77.0 & --    & --
  \\
  \multicolumn{2}{l}{agree \texttt{match\_coll}} & 67.2 & 40.0  & 85.4  & 68.5 & 33.3  & 68.6  & 66.7 & 17.6  & 68.4  & 70.7 & 44.4  & 57.4
  \\
  \multicolumn{2}{l}{agree \texttt{merged}}      & 79.0 & 100.0 & 100.0 & 81.9 & 83.3  & 100.0 & 77.5 & 35.3  & 88.2  & 78.3 & 55.6  & 72.1
  \\
  \multicolumn{2}{l}{agree \texttt{label\_match}}& 50.5 & 40.0  & 54.2  & 58.2 & 33.3  & 41.2  & 62.3 & 11.8  & 60.5  & 65.6 & 33.3  & 47.1
  \\
  \midrule
  \multicolumn{14}{l}{\textit{Label set overlap, $|a\cap b|/|a\cup b|$ (cluster counts)}} \\
  \addlinespace[2pt]
  \multicolumn{2}{l}{\texttt{yes}}     & 96/135 & 5/5 & 48/48 & 108/147 & 6/6 & 51/51 & 116/159 & 6/17 & 71/76 & 150/221 & 6/9 & 109/136 \\
  \multicolumn{2}{l}{\texttt{matched}} & 72/120 & 1/4 & 40/47 &  59/111 & 1/4 & 29/45 &  42/88  & 2/8  & 28/51 &  70/136 & 2/6 &  55/91  \\
  \multicolumn{2}{l}{\texttt{other}}   &  2/37  & 1/4 &  1/8  &  14/62  & 1/5 &  6/22 &  30/74  & 1/7  & 16/31 &  22/76  & 0/2 &  14/45  \\
  \multicolumn{2}{l}{\texttt{none}}    &  0/0   & 0/0 &  0/0  &   1/8   & 0/1 &  0/0  &   9/31  & 0/5  &  8/13 &  14/53  & 2/3 &   9/31  \\
  \midrule
  \multicolumn{14}{l}{\textit{Sample sizes (rows in band)}} \\
  \addlinespace[2pt]
  \multicolumn{2}{l}{$N$ ($a$)}                  & 186 & 56 & 67 & 232 & 67 &  68 & 231 & 62 &  90 & 392 & 54 & 153 \\
  \multicolumn{2}{l}{$N$ ($b$)}                  & 186 & 18 & 91 & 232 & 14 & 104 & 231 & 45 & 175 & 392 & 29 & 361 \\
  \multicolumn{2}{l}{$N^{a\cap b}$ paired}       & 186 &  5 & 48 & 232 &  6 &  51 & 231 & 17 &  76 & 392 &  9 & 136 \\
  \bottomrule
  \end{tabular}
\end{strip}

Table~\ref{tab:iaa-all} presents our inter-annotator agreement evaluation results.

\clearpage
\subsection{LLM-as-a-judge Evaluation}
\label{sec:llm-judge}

To evaluate whether the discovery tier can be partially automated, we replicate the human annotation task with an eight-persona LLM panel: Gemma-4-31B-it and Qwen3.5-27B each running the same four personas used in $P_{\text{per}}$. Aggregate $\kappa$ on \texttt{is\_narr} between human-majority and LLM-majority is 0.53--0.64 (Tab.~\ref{tab:iaa-v3-combined}), superficially comparable to human-human $\kappa$ (0.50--0.58). The aggregate signal is misleading in two ways.

First, the two LLM ensembles agree with each other at $\kappa = 0.76$--$0.83$ on \texttt{is\_narr} and $0.73$--$0.77$ on the merged label space: substantially above either model's $\kappa$ with humans. The judges agree on a shared distribution that is not the human distribution. Second, Gemma's yes-rates exceed both annotators' in all four cells (by 4 to 28 percentage points), and Qwen's do so on PN (by 8 to 24 points) but not on CO, where Qwen lies between the two annotators. The panel therefore tends toward over-recognition of disinformation framings relative to the human baseline, most clearly on PN.

The gap concentrates on the discovery-relevant labels. For \texttt{None} (out-of-domain), the share of human-flagged clusters recovered by the LLM ensemble does not exceed 26\% at any consensus threshold from $L \geq 1$ to $L = 8$ ($L \geq k$: at least $k$ of the 8 LLM personas voted the label) on PN $P_{\text{per-com}}$, and is undefined on CO $P_{\text{per-clr}}$ where no human voted \texttt{None} (Tab.~\ref{tab:iaa-v3-overlap-rel}). For \texttt{Other} (in-domain, no match), recovery rises to 22--57\% at $L \geq 1$ but degrades at stricter thresholds. Even under unanimous LLM consensus ($L = 8$), only 12 of 76 human-flagged \texttt{Other} candidates on PN $P_{\text{per-com}}$ are recovered (16\%); in the reverse direction, $L = 8$ is high-precision (55--100\% of unanimously-flagged clusters are also human-flagged), but such clusters are rare: only 3 to 16 per run (\texttt{Other} and \texttt{None} together) are both unanimously LLM-flagged and human-flagged (Tab.~\ref{tab:iaa-v3-overlap}).

This fits the argument in §\ref{sec:singleton}. If recognizing a narrative in the open-world setting depends on what each annotator has encountered outside the corpus, then two annotators will differ for that reason, and LLM judges set up as in our experiments have no such individual exposure to draw on.

\begin{table*}[t]
\centering\small
\setlength{\tabcolsep}{4pt}
\caption{Per-source label rates and pairwise IAA across the 4 cells. Humans: $a$, $b$. LLM judges: Gemma-4-31B-it (\emph{gemma}) and Qwen3.5-27B (\emph{qwen}), each running the same 4 personas from P$_{\text{per}}$; LLM rates are the mean over the 4 personas. \textit{yes\%}: $P(\text{is\_narr}=\text{yes})$; \textit{match\%}: $P(\text{yes} \wedge \text{label\_match} \in \text{cand}_1{\ldots}_4)$; \textit{other\%}: $P(\text{yes} \wedge \text{Other (in-domain)})$; \textit{none\%}: $P(\text{yes} \wedge \text{None (out-of-domain)})$. $\kappa$ is Cohen on the indicated label space. \textit{majority $\kappa$} = humans-majority vs LLM-majority (or gemma-majority vs qwen-majority), restricted to rows where both groups have a strict majority.}
\label{tab:iaa-v3-combined}
\begin{tabular}{l c c c c }
\toprule
Metric & P$_{\text{per-clr}}$ & P$_{\text{per-com}}$ & P$_{\text{per-clr}}$ & P$_{\text{per-com}}$ \\
$N$    & 186 & 232 & 231 & 392 \\
\midrule
\multicolumn{5}{l}{\textit{yes\% per source}} \\
\quad $a$ & 66.1 & 58.2 & 66.2 & 54.3 \\
\quad $b$ & 58.1 & 51.7 & 52.8 & 40.3 \\
\quad \emph{gemma} (mean over 4 personas) & 70.7 & 62.6 & 79.0 & 67.8 \\
\quad \emph{qwen} (mean over 4 personas) & 61.3 & 54.2 & 74.6 & 64.3 \\
\addlinespace[2pt]
\multicolumn{5}{l}{\textit{match\% per source}} \\
\quad $a$ & 51.6 & 39.7 & 33.3 & 31.1 \\
\quad $b$ & 51.6 & 33.6 & 22.9 & 21.4 \\
\quad \emph{gemma} (mean over 4 personas) & 64.0 & 53.0 & 42.3 & 40.0 \\
\quad \emph{qwen} (mean over 4 personas) & 57.5 & 51.8 & 59.1 & 53.4 \\
\addlinespace[2pt]
\multicolumn{5}{l}{\textit{other\% per source}} \\
\quad $a$ & 14.5 & 17.2 & 20.3 & 12.8 \\
\quad $b$ & 6.5 & 15.5 & 24.7 & 12.2 \\
\quad \emph{gemma} (mean over 4 personas) & 5.1 & 7.4 & 32.0 & 20.8 \\
\quad \emph{qwen} (mean over 4 personas) & 3.4 & 2.4 & 14.8 & 8.8 \\
\addlinespace[2pt]
\multicolumn{5}{l}{\textit{none\% per source}} \\
\quad $a$ & 0.0 & 1.3 & 12.1 & 10.5 \\
\quad $b$ & 0.0 & 2.6 & 5.2 & 6.6 \\
\quad \emph{gemma} (mean over 4 personas) & 1.6 & 2.2 & 4.7 & 7.0 \\
\quad \emph{qwen} (mean over 4 personas) & 0.4 & 0.0 & 0.6 & 2.0 \\
\addlinespace[2pt]
\midrule
\multicolumn{5}{l}{\textit{IAA — Cohen $\kappa$}} \\
\quad $a$ vs $b$: $\kappa$ \texttt{is\_narr} & 0.497 & 0.564 & 0.554 & 0.578 \\
\quad $a$ vs $b$: $\kappa$ \texttt{merged} & 0.558 & 0.651 & 0.610 & 0.621 \\
\quad $a$ vs $b$: $\kappa$ \texttt{match\_coll} & 0.437 & 0.507 & 0.528 & 0.540 \\
\quad humans-maj vs gemma-maj: $\kappa$ \texttt{is\_narr} & 0.598 & 0.594 & 0.527 & 0.592 \\
\quad humans-maj vs gemma-maj: $\kappa$ \texttt{merged} & 0.658 & 0.681 & 0.467 & 0.547 \\
\quad humans-maj vs qwen-maj:  $\kappa$ \texttt{is\_narr} & 0.587 & 0.643 & 0.544 & 0.617 \\
\quad humans-maj vs qwen-maj:  $\kappa$ \texttt{merged} & 0.664 & 0.740 & 0.445 & 0.536 \\
\quad gemma-maj vs qwen-maj:  $\kappa$ \texttt{is\_narr} & 0.759 & 0.768 & 0.789 & 0.827 \\
\quad gemma-maj vs qwen-maj:  $\kappa$ \texttt{merged} & 0.742 & 0.758 & 0.732 & 0.765 \\
\bottomrule
\end{tabular}
\end{table*}

\begin{table*}[t]
\centering\small
\setlength{\tabcolsep}{4pt}
\caption{Overlap counts between human votes and LLM votes for the two non-taxonomy labels: \emph{None (out-of-domain)} (top) and \emph{Other (in-domain, no match)} (bottom). Humans: $a$, $b$ (both annotators). LLMs: 8 personas (Gemma$\times$4 + Qwen$\times$4). \textit{h-either}: $a$ or $b$ voted the label. \textit{h-both}: both humans voted the label. \textit{L$\geq$$k$}: clusters where $\geq k$ of 8 LLM-personas voted the label. Right block: $|h\cap L|$. Relative agreement \%s in Tab.~\ref{tab:iaa-v3-overlap-rel}.}
\label{tab:iaa-v3-overlap}
\begin{tabular}{l c | c c c c | c c c c | c c c}
\toprule
 & & \multicolumn{4}{c|}{Humans} & \multicolumn{4}{c|}{LLM ($n$ of 8)} & \multicolumn{3}{c}{Overlap (h-either $\cap$ L$\geq$$k$)} \\
\cmidrule(lr){3-6}\cmidrule(lr){7-10}\cmidrule(lr){11-13}
Run & $N$ & $a$ & $b$ & either & both & L$\geq$1 & L$\geq$2 & L$\geq$4 & L$=$8 & L$\geq$1 & L$\geq$4 & L$=$8 \\
\midrule
\multicolumn{13}{l}{\textit{Label = \emph{None (out-of-domain)}}} \\
\addlinespace[2pt]
CO: P$_{\text{per-clr}}$ & 186 & 0 & 0 & 0 & 0 & 4 & 3 & 2 & 0 & --- & --- & --- \\
CO: P$_{\text{per-com}}$ & 232 & 3 & 6 & 8 & 1 & 8 & 7 & 2 & 0 & 1 & 0 & --- \\
PN: P$_{\text{per-clr}}$  & 231 & 28 & 12 & 31 & 9 & 15 & 14 & 8 & 0 & 6 & 5 & --- \\
PN: P$_{\text{per-com}}$ & 392 & 41 & 26 & 53 & 14 & 35 & 30 & 26 & 5 & 14 & 11 & 4 \\
\midrule
\multicolumn{13}{l}{\textit{Label = \emph{Other (in-domain, no match)}}} \\
\addlinespace[2pt]
CO: P$_{\text{per-clr}}$  & 186 & 27 & 12 & 37 & 2 & 15 & 12 & 9 & 3 & 8 & 5 & 3 \\
CO: P$_{\text{per-com}}$ & 232 & 40 & 36 & 62 & 14 & 24 & 23 & 14 & 3 & 16 & 11 & 3 \\
PN: P$_{\text{per-clr}}$  & 231 & 47 & 57 & 74 & 30 & 83 & 78 & 69 & 27 & 42 & 35 & 15 \\
PN: P$_{\text{per-com}}$ & 392 & 50 & 48 & 76 & 22 & 103 & 99 & 76 & 22 & 41 & 36 & 12 \\
\bottomrule
\end{tabular}
\end{table*}

\begin{table*}[t]
\centering\small
\setlength{\tabcolsep}{4pt}
\caption{Relative overlap between humans and the LLM ensemble for the same comparisons as Tab.~\ref{tab:iaa-v3-overlap}, reported as percentages with an explicit denominator. \textbf{h\,$\to$\,L}: $|h\cap L|/|h$-either$|$ — share of human-flagged clusters that the LLM ensemble also flagged. \textbf{L\,$\to$\,h}: $|h\cap L|/|L\geq k|$ — share of LLM-flagged clusters that the humans also flagged. \textbf{J}: Jaccard $=|h\cap L|/|h\cup L|$, symmetric. `---' marks rows where the relevant denominator is 0.}
\label{tab:iaa-v3-overlap-rel}
\begin{tabular}{l c | ccc | ccc | ccc}
\toprule
 & & \multicolumn{3}{c|}{L$\geq$1} & \multicolumn{3}{c|}{L$\geq$4} & \multicolumn{3}{c}{L$=$8} \\
\cmidrule(lr){3-5}\cmidrule(lr){6-8}\cmidrule(lr){9-11}
Run & Label & h$\to$L & L$\to$h & J & h$\to$L & L$\to$h & J & h$\to$L & L$\to$h & J \\
\midrule
CO: P$_{\text{per-clr}}$ & \emph{None} & --- & 0\% & 0\% & --- & 0\% & 0\% & --- & --- & --- \\
CO: P$_{\text{per-com}}$ & \emph{None} & 12\% & 12\% & 7\% & 0\% & 0\% & 0\% & --- & --- & --- \\
PN: P$_{\text{per-clr}}$ & \emph{None} & 19\% & 40\% & 15\% & 16\% & 62\% & 15\% & --- & --- & --- \\
PN: P$_{\text{per-com}}$ & \emph{None} & 26\% & 40\% & 19\% & 21\% & 42\% & 16\% & 8\% & 80\% & 7\% \\
\midrule
CO: P$_{\text{per-clr}}$ & \emph{Other} & 22\% & 53\% & 18\% & 14\% & 56\% & 12\% & 8\% & 100\% & 8\% \\
CO: P$_{\text{per-com}}$ & \emph{Other} & 26\% & 67\% & 23\% & 18\% & 79\% & 17\% & 5\% & 100\% & 5\% \\
PN: P$_{\text{per-clr}}$ & \emph{Other} & 57\% & 51\% & 37\% & 47\% & 51\% & 32\% & 20\% & 56\% & 17\% \\
PN: P$_{\text{per-com}}$ & \emph{Other} & 54\% & 40\% & 30\% & 47\% & 47\% & 31\% & 16\% & 55\% & 14\% \\
\bottomrule
\end{tabular}
\end{table*}

\clearpage
\subsection{Human Validation: Guidelines}

\label{app:annotation-guidelines}

This appendix documents the guidelines provided to annotators. Wording is preserved and only formatting has been adjusted. One imprecision should be flagged: the guidelines use ``framing'' informally, as a near-synonym for how a claim is presented, but the term can also be read as invoking framing analysis as a distinct task. A revised protocol should replace it with the operationalization stated in §\ref{sec:relatedwork} (a superclaim: a normalized declarative assertion which a reader could agree or disagree with and under which multiple texts can be grouped) and avoid ``framing'' as a technical term. The guidelines also do not state a definition of ``narrative'' itself. Annotators were instead grounded in examples and in the retrieved labels from the expert-curated taxonomies each candidate was shown alongside, so the effective question was whether a candidate is of the same kind as those expert-drafted labels. We treat this as a limitation of the pilot protocol (see Limitations).

\subsection{Task Overview}

Annotators evaluate automated methods that extract narrative descriptions from text corpora. Each row contains a \emph{superclaim}: a short description representing the core message of a group of related texts. For each superclaim, the annotator makes two judgments:
\begin{enumerate}
    \item Is this a disinformation narrative? (column \texttt{is\_disinfo\_narrative})
    \item If yes, does it match any of the provided narrative labels? (column \texttt{label\_match})
\end{enumerate}
The annotator also rates confidence in the match (1--5) and leaves notes when required.

\subsubsection{Column: \texttt{is\_disinfo\_narrative}}

\paragraph{Question:} Does this superclaim describe a narrative or framing that could plausibly function as a disinformation narrative in the wild?

\paragraph{Options:}
\begin{itemize}
    \item \textbf{Yes}: the framing could function as a disinformation narrative.
    \item \textbf{No}: incoherent, off-topic, neutral reporting, factual statement, or garbled text.
     \item \textbf{Unclear}: the annotator genuinely cannot decide after rereading. Not a default; used when ambiguity is real.
\end{itemize}

\paragraph{What counts as disinformation:} A narrative or framing that purposefully promotes false information, e.g., by distorting facts, attributing motives falsely, amplifying doubt, or framing events to push an agenda. The claim need not be outright false; misleading by omission counts.

\paragraph{Yes examples:}
\begin{itemize}
    \item ``CO\textsubscript{2} is plant food, not a pollutant'' (contradicts scientific consensus).
    \item ``Scientists manipulate data for funding'' (bad-faith attribution).
    \item ``The West provoked the war in Ukraine'' (inverted framing).
\end{itemize}

\paragraph{No examples:}
\begin{itemize}
    \item Factually accurate statements presented straightforwardly.
    \item Neutral news reporting or unrelated content (sports, weather, administrative).
    \item Incoherent text, word salad, garbled fragments.
\end{itemize}

\paragraph{Separating views from judgment:} The question is whether the claim \emph{could function} as disinformation, not whether the annotator agrees with it. A well-phrased argument the annotator finds reasonable still counts if it promotes a misleading view; a claim the annotator disagrees with politically is not automatically disinformation. If the annotator catches themselves thinking ``I agree with this so it's not disinformation'' (or the reverse), they should reread focusing on the rhetorical structure.

\subsubsection{Column: \texttt{label\_match}}

Filled in only if \texttt{is\_disinfo\_narrative = Yes}; otherwise left blank. The question: of the four provided narrative labels, which (if any) best describes the superclaim?

\paragraph{Dataset topics:} Superclaims (all in English) relate to the topics covered by the Climate Obstruction dataset (narratives with obstructive messaging on climate, e.g.\ from fossil-fuel lobby groups) and the PolyNarrative dataset (climate denial and disinformation surrounding Russia's war against Ukraine).

\paragraph{Options:} Four candidate labels are shown in randomized order (position carries no signal), plus two residual options:
\begin{itemize}
    \item \emph{One of the four candidates}: the candidate adequately captures the superclaim's core message.
    \item \textbf{Other} (in-domain, no match): the superclaim fits the general topics of the datasets, but none of the four candidates match it well.
    \item \textbf{None} (out-of-domain): the superclaim is a disinformation narrative but falls outside the combined scope of the dataset topics.
\end{itemize}
All three outcomes are valid; the task is to identify which situation each row falls into, not to force a candidate match. We expect annotators to pick a candidate when the system retrieved a good one, \textbf{Other} when it did not, and \textbf{None} when the narrative lies on a topic outside the datasets.

\paragraph{How to pick:} A match means the candidate captures the same kind of core message, even with different wording. Stance matters: ``renewables cannot meet demand'' does not match ``renewables are growing rapidly.'' Topic match alone is not enough: ``scientists manipulate data'' does not match ``models are unreliable'' (same topic, different rhetorical move). If two candidates match equally, the annotator picks either and notes the tie. When no candidate fits, the annotator errs toward \textbf{Other}/\textbf{None} rather than forcing a match: a false match is worse than a missed match.

\paragraph{Other vs.\ None:} Notes are required for both.

\begin{center}
\small
\renewcommand{\arraystretch}{1.2}
\begin{tabularx}{\columnwidth}{@{}l X@{}}
\toprule
\textbf{Pick} & \textbf{When} \\
\midrule
\textbf{Other} & Superclaim fits the dataset topics, but the four shown candidates don't match well. \\
\textbf{None}  & Superclaim is a real disinformation narrative but lies outside the dataset topics. \\
\bottomrule
\end{tabularx}
\end{center}

\subsubsection{Column: \texttt{confidence} (1--5)}

Filled in only if \texttt{is\_disinfo\_narrative = Yes}. The annotator rates confidence in the \texttt{match} answer.

\begin{center}
\small
\renewcommand{\arraystretch}{1.2}
\begin{tabularx}{\columnwidth}{@{}c X@{}}
\toprule
\textbf{Value} & \textbf{Meaning} \\
\midrule
1 & Very unsure - had to guess; another reader could plausibly pick differently. \\
2 & Unsure - leaning this way but plausibly wrong. \\
3 & Moderate - seems right; some aspects less clear. \\
4 & Confident - would defend this answer. \\
5 & Certain - unambiguous. \\
\bottomrule
\end{tabularx}
\end{center}

\noindent Low confidence is informative, not a failure: a small number of honest 1s and 2s are more useful than uniform 3s.

\subsubsection{Workflow}

\begin{enumerate}
    \item Read the superclaim.
    \item Set \texttt{is\_disinfo\_narrative}: Yes / No.
    \item If No or Unclear, leave the rest blank. Move on.
    \item If Yes, read the four candidates. Set \texttt{label\_match} to one of them, or to Other / None.
    \item Set \texttt{confidence} (1--5).
    \item Fill notes if \texttt{label\_match} is Other or None (required); optional otherwise.
\end{enumerate}
\clearpage
\section{Discovered Narrative Candidates: Full Listings}
\label{sec:app-discovered}

\begin{table}[H]
\centering
\small
\renewcommand{\arraystretch}{1.15}
\begin{tabularx}{\columnwidth}{@{}X@{}}
\toprule
\textbf{Community \& Resilience} \\
\midrule
Emphasizes how the oil and gas sector contributes to local and national economies through tax revenues, charitable efforts, and support for local businesses. \\
Focuses on the creation and sustainability of jobs by the oil and gas industry. \\
\midrule
\textbf{Green Innovation \& Climate Solutions} \\
\midrule
Highlights efforts to reduce greenhouse gas emissions through internal targets, policy support, voluntary initiatives, and emissions reduction technologies. \\
Promotes `clean' or `green' fossil fuels as part of climate solutions. \\
\midrule
\textbf{Pragmatism / Pragmatic Energy Mix} \\
\midrule
Portrays oil and gas as essential, reliable, affordable, and safe energy sources critical for maintaining power systems. \\
Emphasizes the importance of oil and gas as raw materials for various non-power-related uses and manufactured goods. \\
\midrule
\textbf{Patriotic Energy Mix} \\
\midrule
Stresses how domestic oil and gas production benefits the nation, including energy independence, energy leadership, and the idea of supporting American energy. \\
\bottomrule
\end{tabularx}
\caption{Climate Obstruction Dataset: Narrative Taxonomy.}
\label{tab:narrative_taxonomy_co}
\end{table}

Table~\ref{tab:narrative_taxonomy_co} shows the original CO taxonomy, consisting of 7 narratives grouped into 4 narrative groups. 

\begin{table*}[t!]
\centering
\footnotesize
\caption{CO candidates both annotators marked \textit{None (out-of-domain)}.}\label{tab:co-ood}
\begin{tabular}{@{}r@{~}p{0.40\textwidth}r@{\hskip 1em}r@{~}p{0.40\textwidth}r@{}}
\toprule
id & Narrative Label & \#Src & id & Narrative Label & \#Src \\
\midrule
1 & The Trump administration's rejection of social engineering represents a positive and necessary policy direction. & 4 & & & \\
\bottomrule
\end{tabular}
\end{table*}

\begin{table*}[t!]
\centering
\footnotesize
\caption{CO candidates both annotators marked \textit{Other (in-domain, no match)}. Top block: P$_{\text{per-com}}$ (graph, $n{=}14$); bottom block: P$_{\text{per-clr}}$ (cluster, $n{=}2$).}\label{tab:co-other}
\begin{tabular}{@{}r@{~}p{0.40\textwidth}r@{\hskip 1em}r@{~}p{0.40\textwidth}r@{}}
\toprule
\# & superclaim & \#Src & \# & superclaim & \#Src \\
\midrule
\multicolumn{6}{@{}l}{\textit{Graph communities}} \\[2pt]
1 & Market-driven energy production is superior to government regulation and restrictive environmental mandates. & 99 & 8 & Renewable energy is not truly sustainable because it remains fundamentally dependent on fossil fuels. & 27 \\
2 & U.S. domestic energy production is environmentally and ethically superior to that of foreign competitors. & 99 & 9 & Public-private partnerships between energy corporations and public institutions are essential for community safety and environmental conservation. & 25 \\
3 & Energy sector workers must use their voting power to elect leaders who favor deregulation. & 49 & 10 & Major fossil fuel companies possess the leadership and expertise necessary to solve the climate crisis. & 16 \\
4 & Corporate energy entities like ConocoPhillips are responsible actors committed to global sustainability and pandemic relief. & 34 & 11 & U.S. energy production is a cleaner and more sustainable alternative to that of global competitors. & 2 \\
5 & Strategic investments in grid infrastructure and market-driven energy policies are required for long-term societal stability. & 34 & 12\ & Industry-led coalitions are the most effective mechanism for scaling and implementing environmental solutions. & 1 \\
6 & Free-market principles and domestic oil production are essential for national energy security and economic growth. & 28 & 13\ & Market-led innovation is the most effective mechanism for developing cleaner and more reliable energy systems. & 1 \\
7 & The Dakota Access Pipeline is a safe, legally compliant, and ethically justified infrastructure project. & 28 & 14 & The proposed oil tax increase is harmful and should be defeated. & 1 \\
\midrule
\multicolumn{6}{@{}l}{\textit{Claim clusters}} \\[2pt]
1 & Organized advocacy is necessary to correct negative public perceptions of the oil and gas workforce. & 27 & 2 & Opposition to the Permian Highway Pipeline is driven by the spread of misinformation. & 26 \\
\bottomrule
\end{tabular}
\end{table*}

\begin{table*}[t!]
\centering
\footnotesize
\caption{PN candidates both annotators marked \textit{None (out-of-domain)}, cluster- and graph-based pipelines.}\label{tab:pn-ood}
\begin{tabular}{@{}r@{~}p{0.40\textwidth}r@{\hskip 1em}r@{~}p{0.40\textwidth}r@{}}
\toprule
id & Narrative Label & \#Src & id & Narrative Label & \#Src \\
\midrule
\multicolumn{6}{@{}l}{\textit{Graph communities}} \\[2pt]
1 & The United States employs terrorism, double standards, and illegal acts to exert geopolitical control. & 203 & 9\ & Big Tech companies and global censors deliberately suppress conservative content to control the public narrative. & 1 \\
2 & The Biden administration's domestic and foreign policies prioritize ideological agendas over U.S. national security and citizen welfare. & 111 & 10\ & Current geopolitical events are the fulfillment of biblical prophecies that signal an urgent need for spiritual salvation. & 1 \\
3 & Western governments and institutions are systematically eroding civil liberties and democratic norms under the guise of security and crisis management. & 80 & 11\ & Former intelligence officials should be held legally accountable for coordinating with the Biden campaign to influence the 2020 election. & 1 \\
4 & National security depends on total self-reliance in energy and military production to eliminate foreign strategic vulnerabilities. & 59 & 12\ & The 2020 US presidential election was fraudulent. & 1 \\
5 & Neoliberal capitalism and political elites exploit global disasters to facilitate wealth transfer and systemic control. & 59 & 13\ & The monarchy is systematically eroding traditional religious practices to implement a synthetic, universal belief system. & 1 \\
6 & Poland is pursuing an expansionist and opportunistic geopolitical agenda in Eastern Europe. & 32 & 14\ & The Roman Catholic Church is a malevolent organization conspiring to dismantle American liberties and establish a satanic global order. & 1 \\
7 & Western governments are engaging in systemic censorship and the erasure of shared historical legacies. & 30 &  &  &  \\
8 & Viktor Orbán's long-term geopolitical strategy will prevail over the short-sightedness of current Western leaders. & 27 &  &  &  \\
\midrule
\multicolumn{6}{@{}l}{\textit{Claim clusters}} \\[2pt]
1 & The European Union is using political coercion and interference to force Hungary into submission. & 34 & 6 & The Biden family and Democratic leadership operated a systemic criminal enterprise involving foreign bribery and money laundering. & 13 \\
2 & The rise of far-right political movements is a necessary correction to the EU's failed trajectory. & 26 & 7 & U.S. government institutions and intelligence agencies have been weaponized to protect the Biden administration and sabotage Donald Trump. & 13 \\
3 & Pro-Western popular uprisings are foreign-imposed interventions that lead to national destruction and instability. & 25 & 8 & Germany's pursuit of green energy and the phase-out of nuclear power have crippled its economic security. & 9 \\
4 & The Bulgarian government and pro-Western elites are agents of foreign interests who betray the nation. & 17 & 9 & The German government lacks true national sovereignty and is subservient to US, EU, and NATO interests. & 9 \\
5 & Corrupt financial networks involving the Biden family funded international terrorism and criminal activities. & 13 & & & \\
\bottomrule
\end{tabular}
\end{table*}

\begin{table*}[t!]
\centering
\footnotesize
\caption{PN candidates both annotators marked \textit{Other (in-domain, no match)}, graph-based pipeline.}\label{tab:pn-other-graph}
\begin{tabular}{@{}r@{~}p{0.40\textwidth}r@{\hskip 1em}r@{~}p{0.40\textwidth}r@{}}
\toprule
id & Narrative Label & \#Src & id & Narrative Label & \#Src \\
\midrule
1 & There are conflicting perspectives regarding the validity, cause, and controllability of global climate change. & 328 & 12 & The Bulgarian government is compromising national security and legitimacy by supporting Ukraine. & 50 \\
2 & NATO is an offensive instrument of US hegemony rather than a defensive alliance. & 238 & 13 & The conflict in Ukraine is characterized by unsustainable military costs and escalating foreign proxy involvement. & 30 \\
3 & Western military and financial support prolongs the conflict without altering the strategic outcome. & 154 & 14 & Russian air defense systems are effectively neutralizing Ukrainian aerial and maritime attacks to protect domestic territory. & 14 \\
4 & There is a profound conflict regarding the feasibility, morality, and efficacy of the global energy transition. & 144 & 15 & Ukraine employs sabotage and nationalist ideologies to secure geopolitical leverage and maintain military manpower. & 14 \\
5 & Environmental policies are strategic tools used to seize control of food systems and human mobility. & 116 & 16 & A warming limit of 1.7°C is a more realistic and cost-effective target than 1.5°C. & 1 \\
6 & U.S. interference in Ukraine and aggressive containment of Russia have destabilized global security and provoked conflict. & 111 & 17\ & Climate activism tactics are destructive stunts rather than legitimate forms of political expression. & 1 \\
7 & Western governments and the military-industrial complex are exploiting the war for financial and strategic gain. & 106 & 18\ & NATO air defense systems are insufficient to counter Russian heavy glide bombs. & 1 \\
8 & Donald Trump's return to the presidency is the primary catalyst for ending the Russia-Ukraine war. & 100 & 19\ & Reports of widespread coral reef destruction and mass bleaching are exaggerated and contradicted by recovery evidence. & 1 \\
9 & The current international order is characterized by Western hypocrisy and the strategic rehabilitation of far-right ideologies to combat Russia. & 80 & 20\ & The current Ukrainian political order was established through an illegal coup d'état. & 1 \\
10 & Aggressive environmental transitions threaten national security, food stability, and industrial competitiveness. & 61 & 21\ & The seizure of Russian assets to fund Ukraine is an illegal act that undermines global legal stability. & 1 \\
11 & Western military and diplomatic escalation in Ukraine increases the risk of a direct global conflict. & 61 & 22\ & The US national security apparatus weaponizes anti-Russian propaganda to manipulate domestic political outcomes. & 1 \\
\bottomrule
\end{tabular}
\end{table*}

\begin{table*}[t!]
\centering
\footnotesize
\caption{PN candidates both annotators marked \textit{Other (in-domain, no match)}, cluster-based pipeline.}\label{tab:pn-other-cluster}
\begin{tabular}{@{}r@{~}p{0.40\textwidth}r@{\hskip 1em}r@{~}p{0.40\textwidth}r@{}}
\toprule
id & Narrative Label & \#Src & id & Narrative Label & \#Src \\
\midrule
1 & Russia is strategically justified in using nuclear weapons to ensure its sovereignty against Western aggression. & 95 & 16 & The US military-industrial complex and corporate interests are driving the war for financial profit. & 24 \\
2 & International peace conferences are futile and illegitimate because they exclude Russia from the dialogue. & 68 & 17 & The international community must formally recognize and hold the Ukrainian regime accountable for state terrorism. & 22 \\
3 & The Ukrainian military is intentionally targeting Russian civilians and infrastructure to compensate for battlefield failures. & 66 & 18 & Ukraine collaborates with international terrorist organizations and foreign powers to destabilize global security. & 22 \\
4 & Peace in Ukraine is contingent upon Ukraine accepting Russia's territorial annexations and adopting a neutral, demilitarized status. & 65 & 19 & Russia's strategic endurance and operational superiority will force Ukraine to accept the Kremlin's terms. & 21 \\
5 & The sustainability of Western military support is undermined by depleted reserves and industrial limitations. & 47 & 20 & Russia's superior industrial capacity and resource mobilization ensure its inevitable military victory in Ukraine. & 21 \\
6 & The U.S. and Western allies are escalating the conflict by removing restrictions on long-range weaponry. & 45 & 21 & The United States is directly responsible for terrorist attacks on Russian civilians through its operational control of Ukrainian strikes. & 21 \\
7 & Western intervention escalates the conflict and risks a direct hot war between NATO and Russia. & 41 & 22 & Russia and Ukraine both utilize energy supplies and transit as geopolitical weapons to destabilize Europe. & 18 \\
8 & A sustainable peace requires direct high-level negotiations between the United States and Russia. & 37 & 23 & Ukrainian intelligence employs coercion, bribery, and psychological manipulation to recruit Russian citizens and military personnel. & 18 \\
9 & Hungary and Slovakia must reject military support for Ukraine in favor of a negotiated peace. & 34 & 24 & Ukraine is engaging in nuclear terrorism by targeting nuclear facilities and developing radiological weapons. & 17 \\
10 & The Russian state and its security apparatus are effectively neutralizing foreign threats and maintaining domestic order. & 34 & 25 & The United States must cease its support for Ukraine to force a negotiated diplomatic settlement with Russia. & 16 \\
11 & Russia maintains a superior defensive posture that successfully protects its territory from Ukrainian aerial and maritime attacks. & 32 & 26 & The United States should terminate financial and military aid to Ukraine to stop the waste of national resources. & 16 \\
12 & NATO member states lack the operational readiness and strategic will to repel a large-scale Russian invasion. & 25 & 27 & The destruction of the Nord Stream pipelines was a predetermined act of international terrorism threatening Eurasian security. & 11 \\
13 & NATO's military technology and air defenses are fundamentally incapable of countering modern Russian weaponry. & 25 & 28 & The United States government orchestrated the sabotage of the Nord Stream pipelines to eliminate Russian energy advantages. & 11 \\
14 & Direct military confrontation between NATO and Russia would inevitably trigger a catastrophic global nuclear war. & 24 & 29 & Ukraine and its Western allies are orchestrating chemical weapon provocations to frame Russia and secure strategic victory. & 10 \\
15 & The United States is exploiting its European allies by shifting the financial risks and burdens of the war onto them. & 24 & 30 & The Ukrainian government and Western entities collaborate to conduct illegal medical experiments and organ harvesting. & 5 \\
\bottomrule
\end{tabular}
\end{table*}

None-annotated cases are documented in Table~\ref{tab:co-ood}. Only one case had both annotators agreeing on this label, and after curation we consider it truly out of domain. 
In Table~\ref{tab:co-other} a pattern is noticeable among the graph rows 1, 6, 9, 12, 13 and 14: all six can be grouped together at another level of abstraction, since they all argue against potential phenomena of state regulation and for letting the market decide. All six (and their source texts) can be grouped into the narrative proposed in §\ref{sec:discovered}.

The table rows are sorted by source count. \#Src is the number of source documents. A row with \#Src $=1$ is not necessarily a singleton: several claims extracted from the same document can form a community with more than one node, so the singleton counts of Tab.~\ref{tab:singleton-counts} differ from the number of rows with \#Src $=1$. \textit{None} marks content outside the corpus's topical scope, and \textit{Other} content that fits the topic but matches no candidate label from the reference taxonomies. 
The corresponding PolyNarrative candidates are listed separately by pipeline family: Table~\ref{tab:pn-ood} for candidates both annotators marked \textit{None (out-of-domain)}, and Tables~\ref{tab:pn-other-graph} and~\ref{tab:pn-other-cluster} for the graph- and cluster-based \textit{Other (in-domain, no match)} candidates.

All results, including the traces from source texts to their narratives, are published in the supplemental material in our code repo.
\end{document}